# How human-robot collaboration impacts construction productivity: an agent-based multi-fidelity modeling approach

Minghui Wu[a,b], Jia-Rui Lin[b,*], Xin-Hao Zhang[b]

[a] Department of Civil and Environmental Engineering, University of Michigan Ann Arbor, Michigan, USA 48109
[b] Department of Civil Engineering, Tsinghua University, Beijing, China 100084

**Abstract**

Though construction robots have drawn attention in research and practice for decades, human-robot collaboration (HRC) remains important to conduct complex construction tasks. Considering its complexity and uniqueness, it is still unclear how HRC process will impact construction productivity, which is difficult to handle with conventional methods such as field tests, mathematical modeling and physical simulation approaches. To this end, an agent-based (AB) multi-fidelity modeling approach is introduced to simulate and evaluate how HRC influences construction productivity. A high-fidelity model is first proposed for a scenario with one robot. Then, a low-fidelity model is established to extract key parameters that capture the inner relationship among scenarios. The multi-fidelity models work together to simulate complex scenarios. Based on the simulation model, the twofold influence of HRC on productivity, namely the supplement strategy on the worker side, and the design for proactive interaction on the robot side, are fully investigated. Experimental results show that: 1) the proposed approach is feasible and flexible for simulation of complex HRC processes, and can cover multiple collaboration and interaction modes; 2) the influence of the supplement strategy is simple when there is only one robot, where lower Check Interval (CI) and higher Supplement Limit (SL) will improve productivity. But the influence becomes much more complicated when there are more robots due to the internal competition among robots for the limited time of workers; 3) the productivity per robot improves when there are more robots and workers, even if the human-robot ratio remains the same; 4) introducing proactive interaction between robots and workers could improve productivity significantly, up to 22% in our experiments, which further depends on the supplement strategy and the human-robot ratio. Overall, this research contributes an integrated approach to simulate and evaluate HRC's impacts on productivity as well as valuable insights on how to optimize HRC for better performance. The proposed approach is also useful for the development of new robots, optimization of HRC process to maximize construction performance and occupational health.

*Corresponding author: lin611@tsinghua.edu.cn, jiarui_lin@foxmail.com

# 1 Introduction

Construction robots have drawn attention in research and practice these years to cover the shortage of conventional construction. The productivity for the entire industry has been declining in recent years and the conventional construction paradigm has reached the technological performance limit[1]. As for workers in the industry, construction tasks are usually of high physical demand, and sometimes are harmful to their health[2]. For example, many masons are facing low back disorder due to heavy workload [3]. Besides, it is also observed that young generations show low interest in the construction sectors[1], which may further aggravate aging and lacking problems of construction workers. Therefore, construction robots have aroused increasing interest in the last 15-20 years[4] and are gradually implemented in construction projects[5] because they can improve productivity while replacing workers from doing heavy duties and dangerous tasks. The application of robots involves nearly every construction-related task, including glazing[6], beam assembly[7], earthmoving[8], concrete wall fabrication[9], etc.

However, the construction industry cannot be fully automated currently even with the aid from robots[4]. As a result, various operations and tasks still need to be fulfilled by human workers and they have to interact with robots. In other words, human-robot collaboration (HRC) is and will be a critical part of the construction process. Thus, it is important to investigate the impact of HRC process on construction productivity. However, considering the high cost of construction robots[10], it is unrealistic to analyze HRC through field tests. On the other hand, due to the multifarious tasks of both workers and robots, the complicated interactions between the two participants, together with the complex environment settings[11], it is very difficult to formulate a mathematical model or use physical simulation to capture the whole process with acceptable accuracy and computation efficiency. As a result, the HRC process in construction is seldom studied so far, which calls for further investigation.

To address this problem, "bottom-up" approaches such as agent-based (AB) modeling are used in this study to model HRC process. AB modeling is a simulation approach to model systems by using virtual agents to imitate the behaviors and interactions of participated individuals[12]. It is suitable for systems whose[13]: 1) problem domain is spatially distributed; 2) subsystems exist in a dynamic environment and 3) interact with each other. Furthermore, simulation models are effective tools to analyze how critical factors are affecting system performance. For example, Sun et al.[14] compared the evacuation performance for multiple building layout scenarios created by adjusting parameters (i.e. number of exits or doorway width) in the AB model, while Jung et al. [15] aimed at finding the optimal lift system combination among alternatives based on simulation models. The impact of HRC process on construction can be investigated quantitatively in a similar way by analyzing the simulation results with different factors.

This research starts with an analysis and review of HRC in construction, which shows specific complexity and uniqueness comparing to other collaboration systems such as human-human collaboration or human-machine collaboration. Then, the bricklaying process is chosen as a typical application in the construction domain. An agent-based (AB) modeling approach is adopted to simulate HRC process in a multi-fidelity way. First, a high-fidelity model is proposed for a simple scenario with only one robot and three workers. This model thoroughly considered the physical location and detailed behaviors of robots and workers. Three human factors, forgetting, muscle fatigue, behavior uncertainty, are incorporated to further improve the model accuracy and stress the difference between robots and workers. Then, to generalize the model to more complex scenarios with multiple robots, three key parameters are identified that capture the inner relationship between scenarios. To extract the values of the parameters, a low-fidelity model is proposed. The model, simplifying the behaviors of agents, is much easier to build and modify. These parameters are finally used in the high-fidelity model to simulate complex scenarios. The simulation model is then validated. Based on the agent-based hybrid model, we propose an in-depth investigation of how HRC process will impact construction productivity, from both the worker side and the robot side. To improve productivity, workers can develop a better strategy to collaborate with robots, and robots can be designed to be more proactive in interaction with workers.

The remainder of the article is organized as follows. Section 2 reviews previous studies related to HRC process in construction. Section 3 introduces the methodology of this research. Section 4 proposes the development of the high-fidelity AB simulation model. Section 5 discusses the multi-fidelity simulation approach to apply the model in complex scenarios. Section 6 validates the simulation model while provides analysis and discussion of how the HRC process affects construction productivity. Section 7 reviews and discusses key points in the development and utilization of the simulation model. Section 8 summarizes the study, discusses the limitations and future developments of this research.

## 2 Related works

This section synthesizes the studies related to this research. Specifically, Section 2.1 discusses the complexity and uniqueness of HRC in construction comparing with other collaboration systems. Section 2.2 reviews the previous methods to investigate HRC process.

## 2.1 Analysis of human-robot collaboration systems in construction

### 2.1.1 Collaboration system classification

Generally, collaboration systems are classified by participants. In construction, collaboration systems between human-human, human-machine, machine-machine and human-robot are common. Human-human and machine-machine collaboration refer to the collaboration among workers and machines respectively. The difference between human-machine collaboration and human-robot collaboration is that human conducts a mission by manipulating a subordinate[11] (i.e. infrared cameras in bridge inspection[16]) while human and robots can work on different tasks side by side, which indicates the higher autonomy and flexibility of robots[11].

Several communication modes (Table 1) and interaction modes (Table 2) are involved in collaboration systems[17]. These modes are originally used to describe the collaboration between humans and robots, but it is believed that they can be generalized into other collaboration systems as well. Agents in Table 1 referred to the participants in collaboration systems (i.e. robots in HRC). For example, if two agents use cellphones to communicate, then this communication mode is classified as Remote Contactless Interaction. If two agents work in the same place and same time but conduct two different tasks on the same item, this interaction mode belongs to Cooperation.

Table 1. Communication modes in collaboration systems[17]

| Communication mode | Description |
| --- | --- |
| Direct physical interaction | One agent's body contact with another in order to perform a task |
| Remote contactless interaction | Agents contact by interfaces (e.g. voice, camera) |
| Teleoperation | On agent directly drive another with interfaces |
| Message exchange | Information is exchanged using digital signals transmitted through physical buttons |

Table 2. Interaction modes in collaboration systems[17]

| Interaction mode | Description | | | |
| --- | --- | --- | --- | --- |
| | Same place | Same time | Same item | Same task |
| Coexistence | √ | √ | × | × |
| Synchronized | √ | × | √ | × |
| Cooperation | √ | √ | √ | × |
| Collaboration | √ | √ | √ | √ |

**2.1.2  Complexity and uniqueness of human-robot collaboration system in construction**

Traditional collaboration systems in construction between human-human, machine-machine and human-machine are involved in simulation scenarios of previous papers. We summarize the simulation scenarios, collaboration types, communication modes and interaction modes of representative works in Table 3 according to the classification in [17]. For example, in the bridge inspection process, human-human collaboration and human-machine collaboration both occur. In terms of human-human collaboration, preparation and inspection are conducted in sequence by technicians, so the interaction mode between two teams of workers belongs to Synchronized mode. Technicians use voice to communicate, which is considered a form of Remote Contactless Interaction. As for human-machine collaboration, during the inspection process, technicians manipulate and move with the devices to fulfill tasks, which implies that the worker and machine are working on the same tasks all the time, therefore it is classified as Collaboration mode. Besides, since technicians need to set and program the devices first, the communication mode is Message exchange.

It can be concluded that traditional collaboration systems only involve one or two simple communication modes. This is reasonable since humans only need to communicate by voice; the machine-machine system as a fully automated system only needs electronic signals to send messages; while machines as passive objects only need to be programmed or manipulated by humans. Furthermore, since machines lack autonomy and flexibility, collaboration systems with machines only support one interaction mode (collaboration), while human-human collaboration systems may involve multiple interaction modes at the same time.

Table 3. Traditional collaboration systems in previous work

| Paper | Scenario | Collaboration type | Communication mode | Interaction mode |
|---|---|---|---|---|
| Seo et al.[18] | Bricklaying | Human-human | RCI | Collaboration & Coexistence |
| Abdelkhalek et al.[16] | Bridge inspection | Human-human | RCI | Synchronized |
| Zhe et al.[19] | Pump maintenance | Human-human | RCI | Synchronized & Collaboration |
| Jabri et al. [20] | Earthmoving | Machine-machine | RCI | Collaboration |

| Yassin et al. [21] | 3D printing reinforced concrete | Machine-machine | RCI | Synchronized |
| Abdelkhalek et al.[16] | Bridge inspection | Human-machine | ME&DPI | Collaboration |
| Jung et al.[15] | Lift system | Human-machine | ME&DPI | Collaboration |

Comparing to the traditional collaboration systems, HRC in construction is complex and unique in the following three aspects. All three aspects are further illustrated in Section 4.

- Simultaneously involving multiple communication and interaction modes: since many robots in construction are not fully automated, they still need to be installed, programmed or manipulated (Message Exchange, Teleoperation) like machines; however, robots have much higher autonomy and independence as a teammate rather than a passive tool[11], which provide possibilities for more communication modes (Remote Contactless Interaction, Direct Physical Interaction) and more interaction modes. For example, for human-robot collaboration in bricklaying, three communication modes and three interaction modes are involved. Detailed information will be given in Section 7.

- Safety requirement: in extremely complicated and distributed environments like construction sites, it will be very hard to set up fences to separate workers and robots[22]. Since human workers are exposed to robots, human-robot collaboration systems need some safety requirements such as safety distance[23]. Safety issues are seldom considered in traditional collaboration systems.

- Different characteristics between participants: the two participants, humans and robots, have different characteristics. For example, human workers are likely to forget or feel tired, but robots are not. This makes HRC unique from other systems as well. For other systems such as human-machine collaboration, when humans are tired, the efficiency of machines will also be affected since the humans and machines are essentially performing the same task and the machine is manipulated by humans.

## 2.2    Previous methods to investigate HRC process

To the best of the author's knowledge, researchers have not paid much attention to the HRC process in construction, possibly due to the still limited application of construction robots. However, in other domains, researchers propose several methods to study HRC process. This section summarizes these previous approaches.

### 2.2.1    Mathematical modeling approach

Mathematical modeling was adopted by many researchers to investigate and optimize HRC process in manufacturing. Fager et al.[24] proposed a working time model of kit preparation with order batching to compare the productivity between manual and robot-supported assembly lines. Total working time contained picking time, sorting time and traveling time. Each type could be calculated using the proposed equations. Bogner et al.[25] built a mathematical model for HRC process in assembly of printed circuit boards, and optimized the task allocation in order to minimize completion time. However, mathematical models could be difficult to apply to more complicated and dynamic systems such as construction. On one hand, the task in kit preparation only involved picking and sorting, and the HRC process only contained one interaction mode (synchronized) and one communication mode (RMI). However, HRC in construction has more complex tasks and multiple modes at the same time. Besides, the traveling route of the operator in [24] was a 7m bidirectional line. In contrast, construction workers usually have to adopt a longer and more flexuous route because of the larger work zone, disordered material stacking and potential hazards[26]. On the other hand, human behavior was considered deterministic in mathematical models, which meant that humans were assumed to carry out tasks strictly[25]. However, as a labor-intensive industry, construction productivity will be heavily affected by human factors. Some human factors, such as muscle fatigue, will affect workers' behavior so they may behave dynamically which is hard to model mathematically.

### 2.2.2 Simulation approach

There are two general simulation approaches in robot-related domains (see Figure 1). One stream of research is the strict physical simulation which aims to imitate the target system as detailedly as possible[27]. A broad range of simulation platforms is classified into this category, including USARSim [28], Gazebo[29] and SAI [30]. These platforms are based on the physical model of the robot's mechanical design and can provide an accurate evaluation of the robot kinematics, dynamics and interaction behavior with the environment[31]. Notably, Brosque et al. [32] were one of the first researchers to apply physical simulation in the construction domain. However, two features of physical simulation prevent it from application in this research. On one hand, the accurate imitation of physical behavior requires considerable computation capacity[32], which is not suitable for macroscopic modeling with multiple robots, materials, and a complicated environment. As a result, the investigation and optimization of the process within physical simulation are confined locally. On the other hand, physical simulation cannot well model the existence of humans, since it cannot model the ergonomic behavior nor the decision-making process of humans. Thus, physical simulation has to adopt a rather indirect way, such as modeling a human as a shape with biomechanical boundary[33, 34], or use the simulation model with only robots to determine their activity sphere which can provide the safety area of humans[32].

The other stream of research is the dynamic simulation, in which significant features of the tasks are abstracted and simplified[27]. In the dynamic simulation, AB simulation is a useful method to represent individual team members and then understand how they affect team performance, which fits in well with the research goal to study HRC, and therefore it is adopted by several researchers. For example, Giachetti et al. [31] used AB simulation to study the human-robot team performance in military scenarios, and Wang et al. [11] studied the importance of trust between humans and robots in HRC by building intelligent agents. The AB simulation in this research is distinguished from the previous researches in four main ways. First, the robot tasks and their interaction with humans in our model are more complicated. In previous works[11] [31], robots mainly focus on moving tasks and do not have much interaction with workers. Second, to deal with the multifarious interaction when there are multiple robots and workers on site, we propose a multi-fidelity simulation method. The high-fidelity model guarantees the accuracy of the simulation, which the low-fidelity model brings extra flexibility. Besides, the AB model we proposed gives attention to the ergonomic behavior of humans, which is important to further stress the difference between humans and workers. Finally, an in-depth investigation of how HRC process affects construction productivity is provided in this research, which is seldom discussed before.

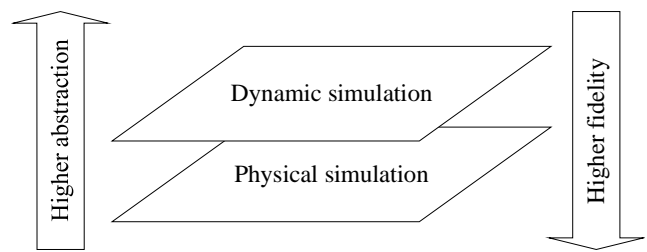

Figure 1. Simulation approaches in the robot domain

## 3 Research methodology

The objective of this research is to model HRC process in construction with high precision and use the proposed model to examine the effect of HRC process on total construction productivity. Considering construction is a rather broad topic, the focus is narrowed down to the bricklaying process in consideration of the following three reasons.

- Bricklaying is one of the most traditional construction processes. Although reinforced concrete (RC) is the dominant structure type nowadays, bricks are still used in infilled walls between two RC columns.

- Bricklaying robots are successfully promoted. The research of bricklaying robots can be traced back to the nineties[35]. Several bricklaying robots, such as SAM100 and Harian X, are derived from these researches and are gradually adopted by companies and projects[36].

- Bricklaying is labor-intensive and carries physical risks[2], which is suitable for the application of construction robots.

To achieve the research objective, a three-step methodology is utilized (Figure 2): high-fidelity model development, generalization of modeling scenarios, and analysis of the influence of HRC process. Each part is introduced as follows.

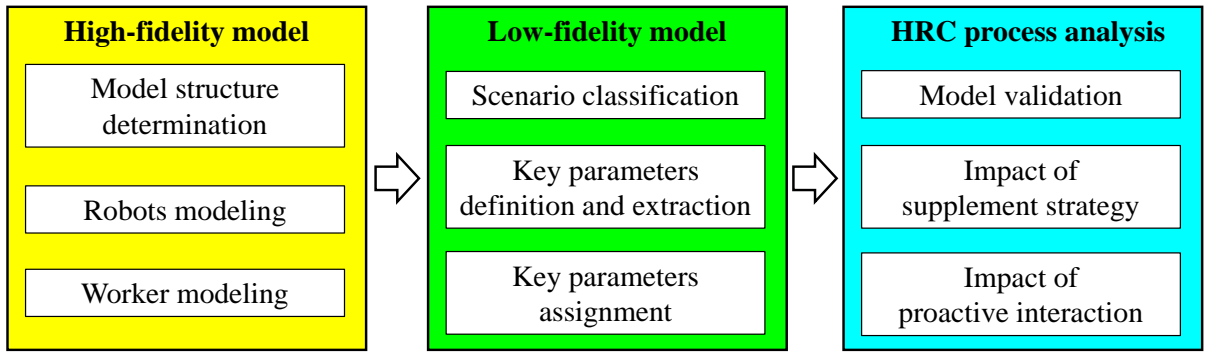

Figure 2．Overview of the research methodology

AnyLogic© (version 8.5.0) is used as a simulation platform for the high-fidelity model. First, types of agents, their relationships and the simulation environment are defined in the model structure. The working hour agent, the robot agent, the worker agent, the brick agent and the sensor agent are incorporated. A real project in Beijing is used as the simulation environment in this research. This section focuses on a simple scenario with one robot and three worker agents in the simulation environment. Then, behavior rules are captured to define the agents' property in the high-fidelity model so that agents can act spontaneously. More specifically, the core of the section is to model the behavior of robots and workers.

To model more complex scenarios, we first propose a scenario classification guideline. In this research, the complex scenarios refer to the scenarios with multiple robots. Then, three key parameters, Occupied Rate (OR), Waiting Time (WT), and Get Checked Interval (GCI), are defined that capture the inner relationship between the simple scenario and the complex ones. A low-fidelity model is proposed subsequently to extract the key parameters. The model, simplifying the behaviors of agents, is much easier to build and modify. These parameters are finally assigned to the high-fidelity model to simulate complex scenarios.

Then, the validation of the multi-fidelity simulation model is tested. Based on the model, the influence of HRC process on construction productivity is analyzed. The influence is two-fold, namely the worker can develop a better strategy to cope with robots while the robots can be designed to be more proactive in the interaction with human workers. We take the strategy of regarding the checking and supplementing frequency and the installation of gravity sensors as examples to thoroughly investigate the twofold influence.

## 4 Development of the high-fidelity model

In this section, based on our previous work[37], the main steps of developing the high-fidelity AB model to simulate HRC process in bricklaying will be introduced and discussed. This model thoroughly considered the physical location and detailed behaviors of robots and workers.

As a highly flexible approach, there are many ways to model the HRC process. Among them, a robot-centered approach is used in this research, making it easier to identify and define the behavior rules of robot agents and worker agents. For example, when dealing with several workers moving a robot to the destination, it is difficult to keep the synchronicity of these workers if they are moving separately. On contrary, from a robot-centered perspective, the robot moves to the destination, and the workers only need to keep walking towards the moving robot, which can keep workers at the same pace.

Furthermore, since AB simulation is a highly abstract approach, the agent in the model does not have physical functions such as laying bricks. Therefore, it is important to find counterparts for all the tasks performed by both robot agents and worker agents to accurately model their behavior. For example, the bricklaying robot needs ten seconds to lay a brick. Although the abstract robot agent cannot really perform this task, if the robot stays fixed at the target place for ten seconds and then a brick is generated, the process will look the same as the real task. Abstractly, tasks can be classified into four categories, and the counterpart for each category is provided according to Table 4. It is easy to see all possible interaction modes in Table 2 can be modeled with the four types of tasks. Besides, AnyLogic© contains a powerful mechanism called message. Agents can send messages to all or several specific agents. This function, together with the four types of tasks, can model the four communication modes in Table 1 as well. Details are illustrated in Table 5. For example, if two agents communicate through Direct Physical Interaction, first they need to move to meet and operate to exchange information. Besides, they need messages to simulate the information they exchanged.

Table 4. Counterparts for different types of actions

| Action classification | Counterpart | Example |
|---|---|---|
| Working Actions | Staying fixed at a given | A robot uses ten seconds to lay a brick: the robot stays at |

| | position for a given time | a target place for ten seconds, then a brick is generated |
|---|---|---|
| Moving Actions | Moving to a given position at a given speed | Workers move a robot: the robot moves to a destination while the workers keep moving to the robot |
| Resting Actions | Staying fixed at a random or given position until needed | A worker rests because of tiredness: the worker stays fixed at a given place until he is able to work again |
| Operating Actions | Both staying fixed at a given position for a given time | Workers use 30 minutes to install a robot: the robot and workers stay fixed at a target place for 30 minutes |

Table 5. Modeling techniques for communication modes

| Communication mode | Required functions |
|---|---|
| Direct Physical Interaction | Moving, operating, message |
| Remote Contactless Interaction | Message |
| Teleoperation | Moving, message |
| Message Exchange | Moving, operating, message |

Following the above-mentioned principles and methods, the overall model structure and all the agents involved in the HRC process are modeled. Their behavior rules are described in this section. Passive agents like brick agents are manipulated by other agents, and have no complex behaviors. Thus, descriptions of them are omitted.

### 4.1 Model structure

Model structure determines the types of agents and their relationships. Initially, four kinds of agents are involved in this model, the working hour agent, the robot agent, the worker agent, the brick agent. Similar to [16], a hierarchical structure is adopted in this research to express the relationship between agents, as shown in Figure 3. The structure contains three layers. The first layer is the workplace where robot agents work together with worker agents to fulfill given tasks by manipulating passive brick agents. Their working time is monitored and captured by a working hour agent, which is also presented on this layer. The simulation environment is defined in this layer, which describes the workplace. In this research, a typical residential project in Beijing is used to provide references to the design of the simulation environment. Several assumptions are made to simplify the original layout of the construction site without losing generalization. One of the buildings in the layout is picked as the construction object. The corner of the site is assumed to be Long-term Store Place to deposit construction materials, such as bricks. A small place near the building is considered as Temporary Store Place for workers' convenience. A rectangle area that envelopes the building is assumed to be Work Zone. Besides, a small rectangle place near Temporary Store Place is set to be Robot Store Place. In the beginning, all bricks are deposited in the Long-term Store Place. The robot is placed in the Robot Store Place, while all the workers are randomly placed in Work Zone. The simplified layout of the simulation environment is shown in Figure 4. It is then imported to the simulation model using real-world plotting scale to ensure reliability.

In the second layer, the behavior rules for the upper four agents are determined using several parameters and state charts. In Figure 3, the behavior rule and the corresponding agent are linked by dashed lines. The relationship between different agents is illustrated by thick lines and explanations. Furthermore, the robot can be designed to interact more proactively with human workers, which enhances collaboration performance. An example of this proactive design is the installation of gravity sensors, which will be further explained in Section 4.5. In the structure of the simulation model, the sensor agent is embedded in the robot agent in the second layer, and its behavior is further defined in the third layer. In this section, the model focuses on the scenario with only one robot agent.

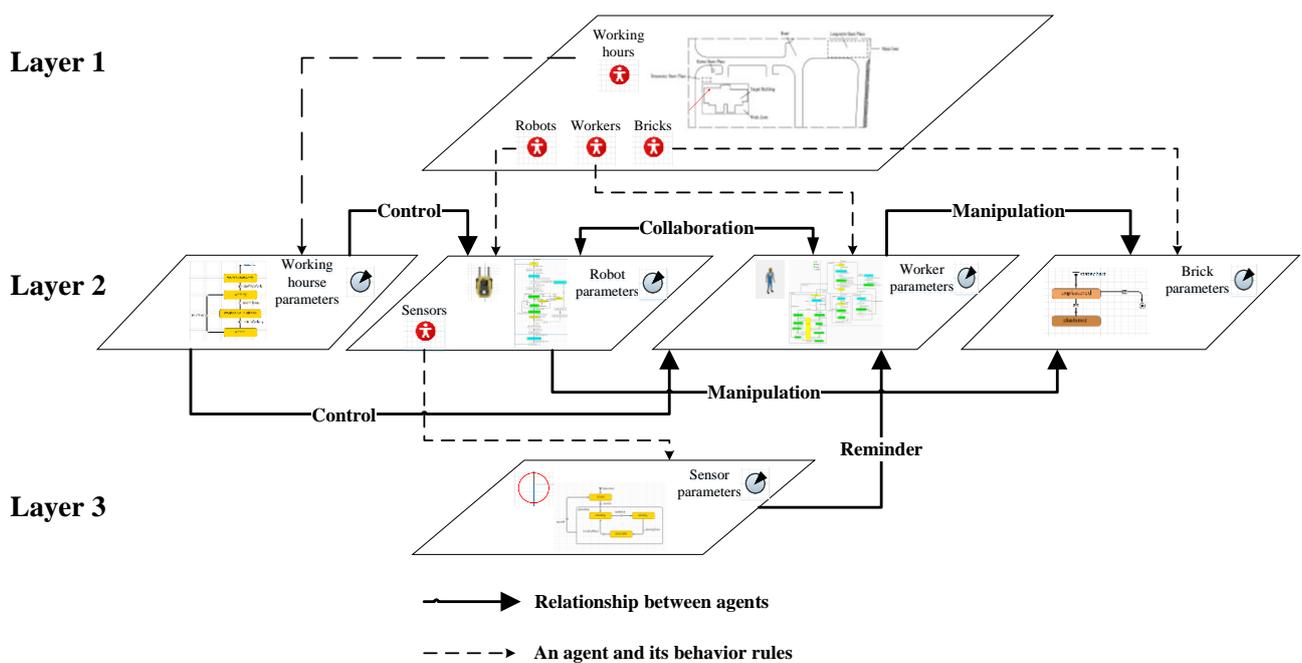

Figure 3. The hierarchical structure of agents

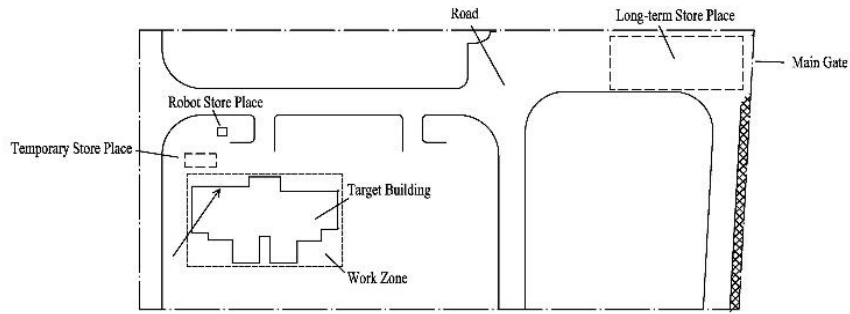

Figure 4. The simulation environment of the model

## 4.2 Modeling working hours

The flowchart of the working hour agent is illustrated in Figure 5. Its interactions with other agents are labeled with arrows. The direction of the arrows indicates the information direction, whether the agent is sending messages or receiving messages. The working hour agent is initialized at "waitToStartWork" state. When the construction period begins, it switches to "working" state and starts the loop. At "working" state, it will inform worker agents to start work. After eight hours, which is a common working period every day, the working hour agent switches to "workingHourFinished" state and informs worker agents. After all the workers get off work, the agent switches to the next state "resting", and will go back to "working state" the next day.

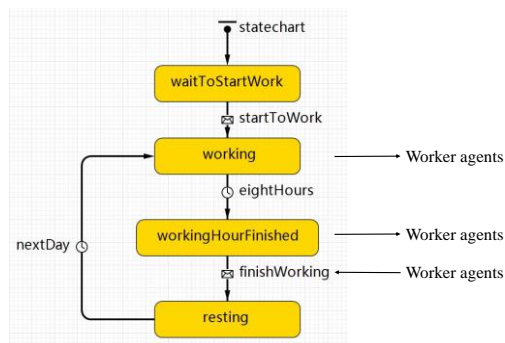

Figure 5. Working hour agent

## 4.3 Modeling robots

### 4.3.1 Agent behaviors

The main job of robot agents is to construct brick walls. Conventionally, brick walls function as infilled walls located between columns. Two dummy nodes, start node and end node, are extracted to represent the endpoints of each wall, as shown in Figure 6.

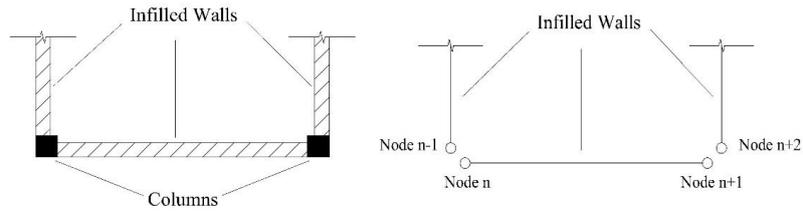

Figure 6. Construction tasks for robot agents

The flowchart of the robot agent and its interaction with other agents are presented in Figure 7. The robot agent is initialized in "robotIdle" state. When workers come and prepare to move the robot, it will be activated and switch to "preparingToMoveToWorkStie" state and will stay at this state for the time need for preparation. This is a typical example of Direct Physical Interaction and Operating Actions. Then, the robot moves to the worksite and informs the workers. After reaching the destination, it will switch to "robotSetting" state, representing workers installing the robot on-site, which is another procedure of Operating Actions. After being installed, the robot will call its collaborators and will receive a message when all collaborators are in position. Then, the robot agent switches to "robotWorking" state and begins to work. After the current wall is finished, or the working hour agent lets the robot go off duty, the robot agent moves to the start point of the current wall and waits for workers. When workers come, if the working hour is not finished and there are unconstructed walls, they will move the robot to the start point of the next wall. Otherwise, they will move the robot back. For the latter occasion, preparation work needs to be done at first, and then the robot moves back to the Robot Store Place. After the robot reaches the destination, the agent will switch back to "robotIdle" state and informs all the workers to go to rest.

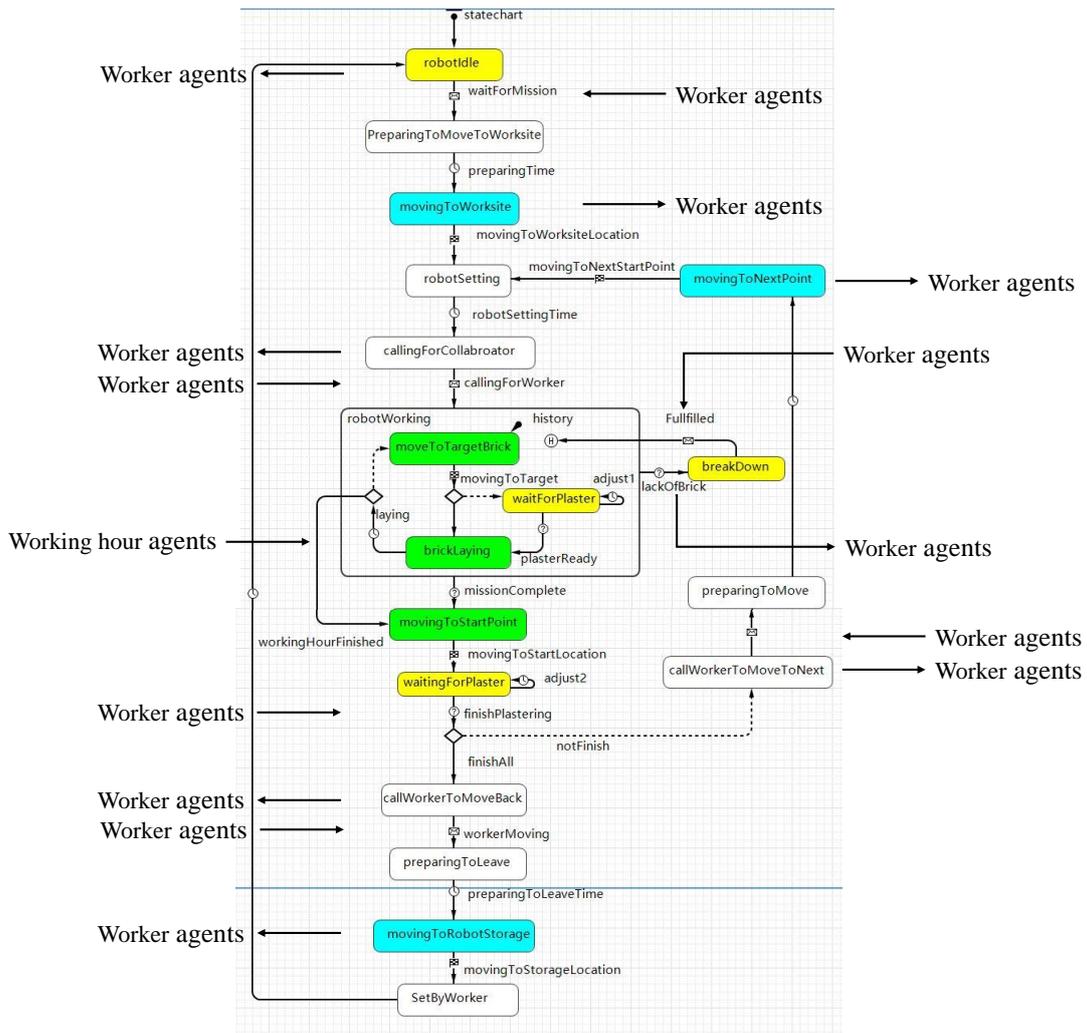

Figure 7. Robot agent

As for the working process, although there are several types of bricklaying robots existing in industry or academia [2, 35, 38-42], most robots share the same process of laying one brick. Therefore, SAM100 is chosen as an example to model the laying process. For SAM100, the whole process can be summarized into the iteration of two procedures, moving to target position and bricklaying. The latter is a generalized process that contains plastering on the surface, grabbing one brick and laying it on the mortar. These two procedures are represented by "moveToTargetBrick" state and "bricklaying" state. When one layer is finished, the robot will return to the start node. For safety reasons, it is assumed that the robot will not start the next layer until the worker responsible for extra mortar removing (introduced in the next section) finishes this layer. Therefore, when the robot agent reaches the target place but the worker is not ready, it will first switch to "waitForPlaster" state and will not switches to "bricklaying" state unless the worker is ready. Since several bricklaying robots have a pipe to deliver mortar[10], it is assumed that the mortar is always sufficient. However, robots will probably still run out of bricks that are stored inside. When this happens, the robot will pass to "breakDown" state and stop operating. When the bricks are supplemented, the robot will start operating again and continue the previous task.

### 4.3.2 Summary of parameters

Parameters of robot agents and their meanings are summarized in Table 6.

Table 6. Parameters and meanings of robot agents

| Parameter | Meaning |
| --- | --- |
| Node Set | The set for all start nodes and end nodes |
| Number of Layers | Number of layers in each wall |
| Brick Laying Duration | Duration for laying one brick |
| Number of Nodes | Number of nodes in the Node Set |
| Moving Velocity in Working | Velocity as robots moving to the target position during working |
| Robot Storage Capacity | The maximum number of bricks that can be stored inside the robot |

## 4.4 Modeling human workers

In the model, workers are responsible for three kinds of tasks in the whole process (Figure 8). First, workers need to carry the robot to target locations. This task is called Robot Moving and Installing (RMI). It is assumed that three workers are needed to perform this task. Second, considering when robots squeeze a brick on the mortar, the robot arm will also push some mortar beyond the below brick's edges, this extra mortar is supposed to be removed by one human worker. This task is called extra mortar removing (EMR). Third, robots may run out of bricks in their storage, so one worker is needed to check periodically and supplement bricks in time[43]. This task is called Brick Supplement (BS). All three worker agents collaborating with a robot agent need to perform the RMI task. Besides, one is responsible for BS (which is called the BS worker), another worker agent is responsible for EMR (which is called the EMR worker), and the third agent can perform other unrelated tasks. The tasks are allocated by the robot agent at "callingForCollaborator" state.

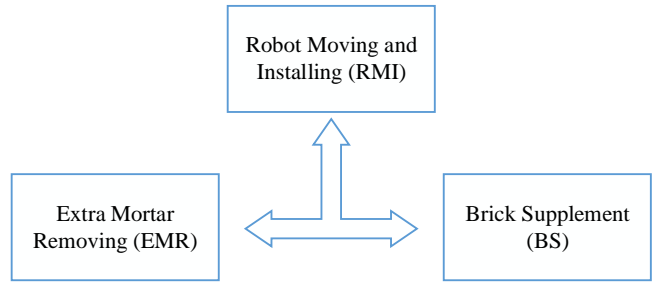

Figure 8. Construction tasks for worker agents

### 4.4.1 The task for robot moving and installing (RMI)

RMI task can be divided into three parts, corresponding to three parts of Figure 9.

(1) As shown in Figure 9 (a), at the very beginning, worker agents are initialized in "workerIdle" state. After informed by the working hour agent, the agents first switch to "movingToRobotStorage", which indicates that they are moving to Robot Store Place. The state switches to "preparing state" when they have arrived. After receiving the message from robot agents indicating the completion of preparation, workers will keep moving to the robot while the robot keeps moving to the destination, which is a typical procedure of Moving Actions. After informed by the robot agents of their arrival, workers switch to "workerSettingRobot" state, which refers to installing the robot on the worksite. The robot agent will inform the workers when the installation is completed, then worker agents will switch to "workerReturn" state, in which the agents will choose a random place to rest as a typical procedure of resting category.

(2) As shown in Figure 9(b), when the robot finishes the bricklaying work for one wall, it will inform workers to move to it. Then, if the working hour is not finished, workers will judge whether there are other unconstructed walls. If so, they will move the robot to the next unfinished wall. Similarly, some preparation work needs to be done. The robot agent will inform the worker agents of the finish of the preparation, then workers will keep moving to the robot while the robot keeps moving to the destination. Their arrival will be informed by the robot agent through message, then the worker agents will switch back to "workerSettingRobot" state in Figure 9(a).

(3) Otherwise, workers will move the robot back to Robot Store Place (Figure 9(c)). This process resembles (2), so it will not be explained in detail. The worker agents will finally switch back to "workerReturn" in Figure 9(a).

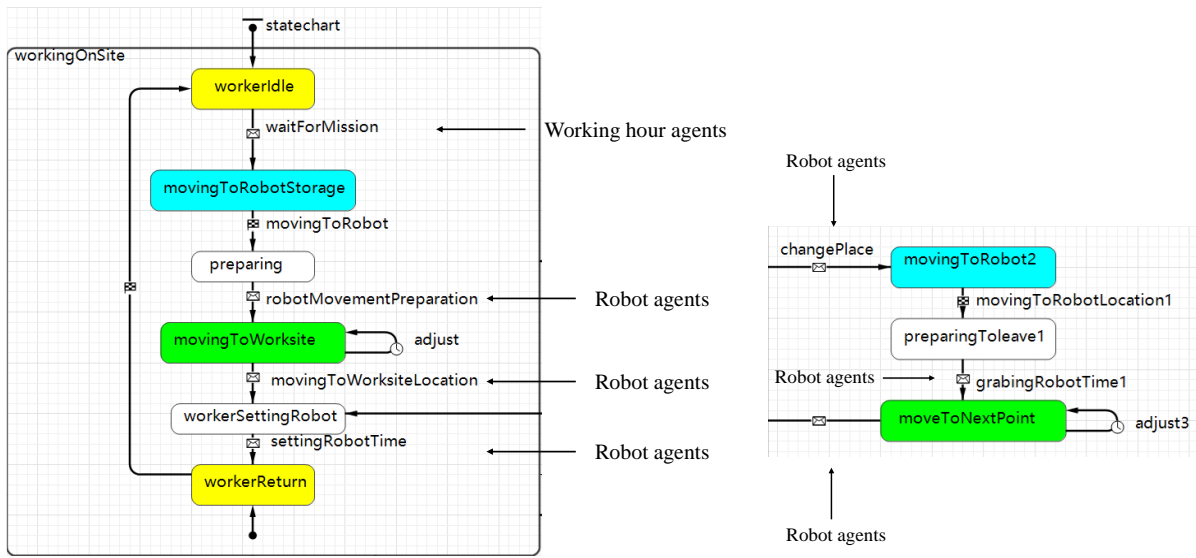

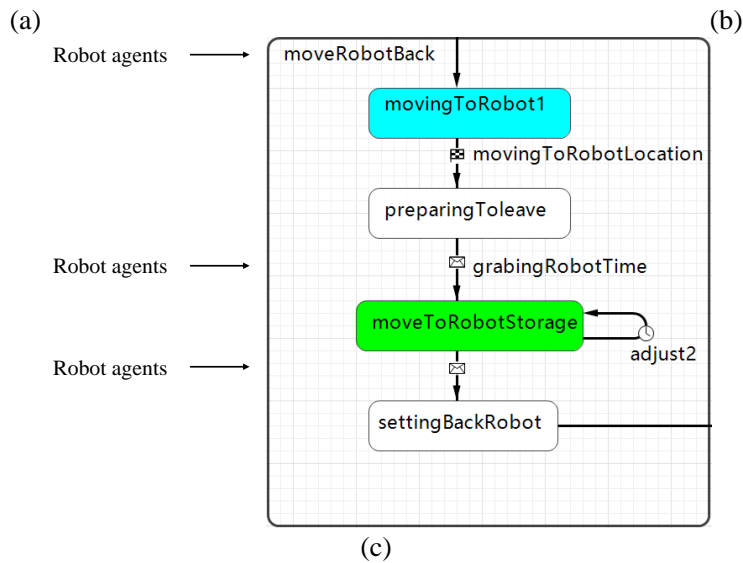

Figure 9. RMI task of worker agents

### 4.4.2 The task for extra mortar removing (EMR)

For EMR tasks and BS tasks (BS tasks will be discussed in Section 4.4.3), three human factors are incorporated.

(1) Forgetting: Because the BS worker may repeat checking many times a day, he or she is possible to forget some checks. To capture this behavior, Remember Probability (RP) is determined by the following equation[44, 45], where CI refers to check interval (the time between two contiguous checks), and $A, B$ refers to parameters.

$$RP = Ae^{-B \times CI} \quad (1)$$

(2) Muscle fatigue: In the bricklaying process, several tasks have physical workloads on masons [46], which leads to muscle fatigue. To address this, we make reference to a dynamic fatigue model proposed by Seo et al. [18]. A worker will take a voluntary rest when his current level of muscle strength is lower than the physical demand of the next work element, and will not perform the task until the former is at least 10%MVC higher than the latter. MVC refers to maximum muscle strength. The relation of current muscle strength and the workload is shown in equation (2), and muscle strength in the recovery process can be explained in equation (3).

$$\frac{F_{cem}(t)}{MVC} = e^{-\frac{F_{load}}{MVC}d} \qquad (2)$$

$$F_{cem}(t_b) = F_{cem}(t_a) + r \times (t_b - t_a) \qquad (3)$$

$F_{cem}(t)$ represents the currently available maximum force at time $t$. $d$ refers to the duration of the task, and $F_{load}$ refers to the average physical demand of a work element. Only four work elements that have physical demand are considered in this research based on Seo et al. (Table 7). Besides, $r$ is the recovery rate, which equals 5%MVC per 1 min when $F_{cem}$ is lower than 90%MVC, and equals 0.3%MVC per 1 min otherwise.

Table 7. Average physical demand for work elements

| Work element | Average physical demand(%MVC) |
| --- | --- |
| Extra mortar removing | 10 |
| Grabbing bricks | 40 |
| Dropping bricks | 40 |
| Adding bricks | 40 |

(3) Behavior uncertainty: Human workers cannot act strictly like robots. The most direct way to model the uncertainty is that the duration workers need to perform a task is not fixed. Therefore, we consider adding triangular distributions with a 20% variance in the parameters related to the workers' task duration to capture their random performance.

The flowchart of the EMR worker agent is given in Figure 10. After he receives the EMR task, the agent will move to the robot. After his arrival, the EMR task begins. The EMR task is the iteration of moving to target brick and removing extra mortar on the brick, corresponding to "moveToTargetBrick" state and "brickPlastering" state. To ensure that the EMR worker has a safe distance from the robot, it is assumed that the distance between the worker agent and the robot agent should be large than the length of 10 bricks. If some workers get too closed to the robot, the robot will stop working. Besides, if the worker is exhausted, the agent will also stop and rest for a moment, which is another typical example of Resting Actions. If the worker agent receives the message from the working hour agent, it will return to "workerReturn" state in Figure 9(a). Note that when the robot runs out of bricks and stops operating, but the BS worker does not show up, the EMR worker will inform him to supplement bricks (i.e. call him by voice or cellphone).

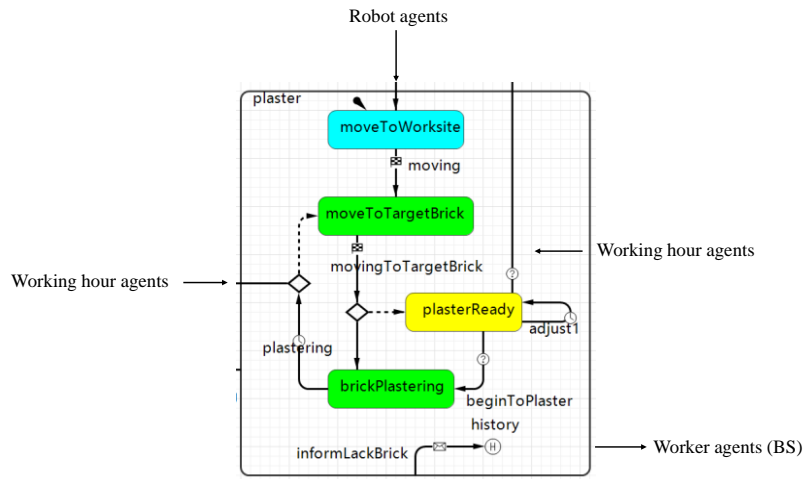

Figure 10. EMR tasks for worker agents

### 4.4.3 The task for brick supplement

The flowchart of the BS worker agent is given in Figure 11. After receiving the mission from the robot agent, the worker agent will switch to "checkIdle" state. There are two ways to begin to perform the BS tasks, which includes

(1): The agent will periodically check the bricks in the robot agent, and this procedure is labeled with a purple rectangle. To improve efficiency, he should not wait until the bricks are used up so that the robot can be supplemented before it runs out of bricks. In other words, if he finds the bricks are less than Supplement Limit (SL), he will start the supplement process directly. The worker agent will go to Temporary Store Place or Long-term Store Place to supplement bricks, corresponding to "movingToBrickTemp" state and "movingToBrickStore" respectively, depending on the number of bricks at Temporary Store Place. If it is larger than the Robot Storage Capacity, he will grab, move and add these bricks to the robot. Otherwise, the BS worker will first transport five times the Robot Storage Capacity bricks from Long-term Storage Place and drop them at the Temporary Storage Place. For each task, if the worker cannot perform the task due to fatigue, the agent will switch to "resting" state to stay until he can work again. After the robot finishes working or the working hour is finished, the BS worker agent will switch back to "workerReturn" state in Figure 9(a).

(2): If the BS worker agent fails to supplement brick in time and causes the robot to use up all the bricks and stop operating, the EMR worker will inform the BS worker to start the supplement process (labeled with a red rectangle). Note that this situation is harmful to the construction productivity, since the robot will stop working for a while. Therefore, it is important to determine the proper frequency for checking and supplementing to avoid this situation.

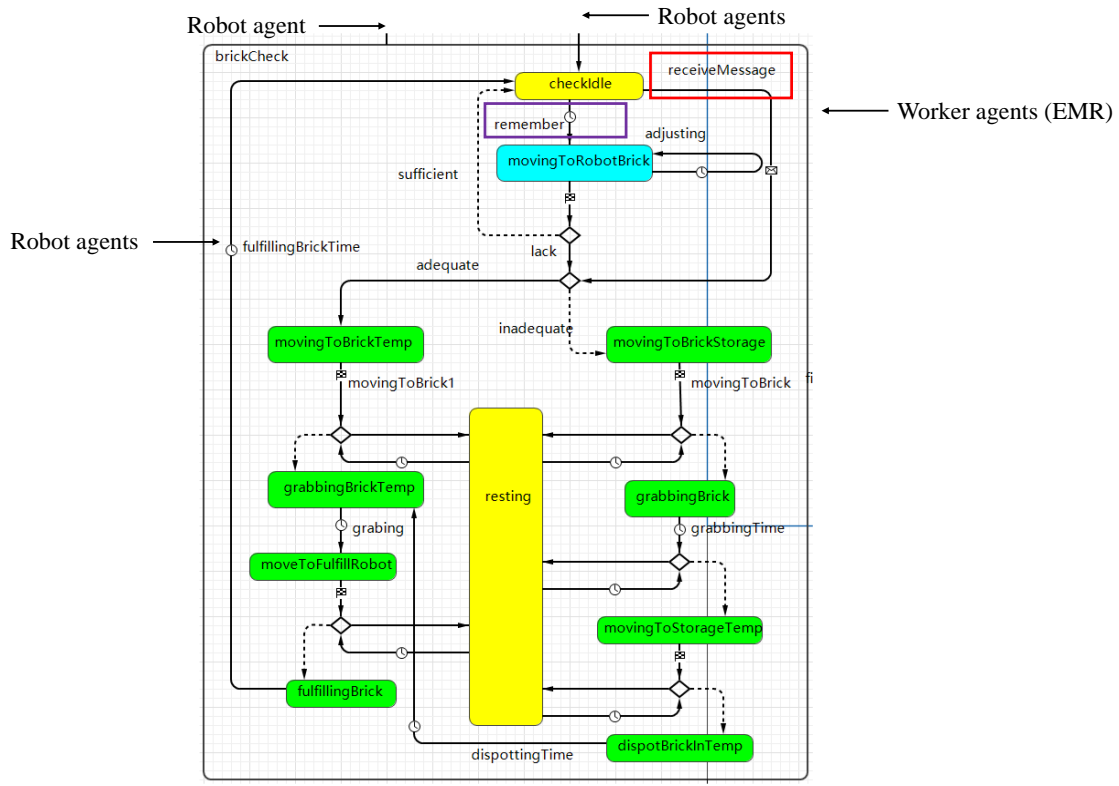

Figure 11. BS tasks for worker agents

Among the three types of workers, the BS worker is considered most important among three workers in the human-robot collaboration process due to the following two reasons: 1) the BS worker interacts more frequently with robots than the other two worker agents, and their interaction is complicated and involves two interaction modes, coexistence and collaboration; 2) The BS process involves two human-determined parameters, SL and CI, which indicates that the BS process can be optimized by developing a better supplement strategy, namely selecting appropriate values for SL and CI. Therefore, the following sections mainly focus on analyzing the BS process. Note that other tasks like the EMR task can be analyzed following the same route.

#### 4.4.4 Summary of parameters

Parameters and meanings of worker agents and their meanings are summarized in Table 8.

Table 8. Parameters of worker agents

| Parameter | Meaning |
| --- | --- |
| Moving Robot Velocity | The moving speed of workers when moving a robot |
| Walking Velocity | Moving speed of workers |
| Mortar Removing Duration | The time for a worker to remove mortar on one brick |

| Supplement Limit (SL) | Under this limit, the worker will start the supplement process |
| --- | --- |
| Check Interval (CI) | The interval between two contiguous check |
| Preparation Duration | The time needed for preparation work |
| Installation Duration | The time needed for robot installation |
| Grabbing Duration | The time needed for grabbing bricks |
| Dropping Duration | The time needed for dropping bricks |
| Supplement Duration | the time needed for supplementing bricks |
| Carrying Velocity | the moving velocity for the BS worker when transporting bricks |
| A, B | Parameters related to forgetting behavior |

## 4.5 Sensor-based proactive interaction

So far, the robot is acting passively in collaboration. For example, in the BS process, the robots just focus on their tasks and only passively wait for workers to come and check. It does not provide any information and guidance to help workers better decide the checking time and frequency. On contrary, the robot can be designed to interact proactively with workers. For example, imagine some gravity sensors are installed that will alert workers when the bricks inside are less than SL. This design can provide useful information about the appropriate action time to workers, so that the workers can supplement bricks in time. This section further develops the model to simulate this new design with gravity sensors. Note that other designs, such as a user interface that can predict the possible time for running out of bricks, can be implemented by summarizing their functions by defining new agents, following the same way as the gravity sensor agent

In the new high-fidelity model, the flowchart of the sensor agent is given in Figure 12. The agent is initialized in "closed" state. When the robot agent begins to work, the sensor agent is activated and switches to "checking" state. At this state, every time the bricks in the robot are insufficient, the agent switches to "alerting" and sends messages to worker agents. After the bricks are sufficient again, the sensor agent will switch back to "checking" state. After the robot finishes its work, the sensor agent will switch to "closed" state again. Note that the sensor will only alert for five seconds and then stop sending alerts. or it will bring too much noise.

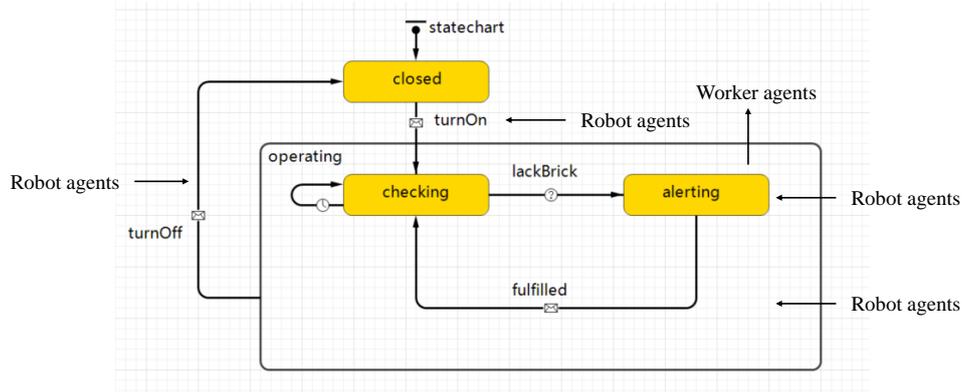

Figure 12. Sensor agent

## 5 Low-fidelity modeling: generalization from simple to complex scenarios

The proposed high-fidelity simulation model focuses on simple scenarios with only one robot agent (with or without gravity sensors) and three worker agents. When there exist more robots or more workers, their duties may have overlap, and the interaction between agents will be complicated, which makes it very difficult and computationally expensive to model the complex scenarios. Instead, we model a sub-collaboration system that only contains one robot, as the average performance of complex scenarios. This generalization approach enables us to model all the complex scenarios with one high-fidelity model. This method has the following two advantages, convenience and flexibility, comparing to building precise simulation models for each complex scenario,

- Convenience: complex scenarios may involve a large number of workers and robots, which poses great difficulty to the modeling procedure. Besides, these agents and their interaction will also cause high computation expenses. In contrast, the generalization approach will not need much extra effort and computation resources.
- Flexibility: Building a precise model for each complex scenario is not flexible. For example, if the number of workers and robots changes, or the robot uses a new type of sensor, the model should be changed significantly or even rebuilt. On the contrary, complex scenarios can be unified to the same simulation model with a proper scenario generalization technique, which makes the model more adaptive.

### 5.1 Scenario classification

First, we give a clear criterion for classifying simple and complex scenarios. As discussed in Section 4.4, the analysis mainly focuses on the BS tasks. Thus, two complex scenarios are considered in this research (Figure 13): 1) Multiple Robots-Single BS Worker scenario, where one BS worker checks and supplements bricks for multiple robots simultaneously; 2) Multiple Robots-Multiple BS Workers, which has several BS workers and several robots on site.

These two scenarios have no restrictions on the installation of gravity sensors. In other words, the existence of gravity sensors does not influence the classification of scenarios, since we have already developed the sensor agent in the high-fidelity model.

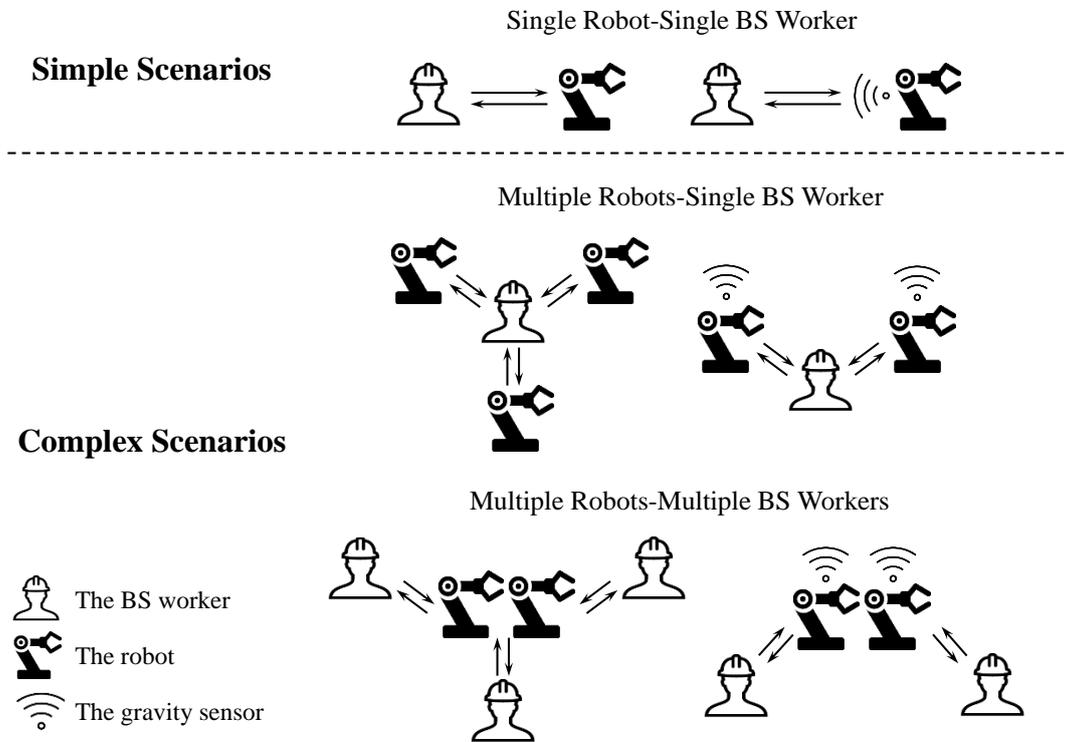

Figure 13. Simple and complex simulation scenarios

## 5.2 Scenario generalization
### 5.2.1 Key parameters to capture the relationship between complex and simple scenarios

In the simple scenario, the interaction between the robot and the BS worker is summarized as follows. If there are no gravity sensors, the BS worker periodically checks the robot; the robot (the EMR worker beside the robot) informs the BS worker when it runs out of bricks. If the sensors are installed, the BS worker can also get notified by the sensors.

For the Multiple Robots-Single BS Worker scenario, in the robot's perspective, it is still checked periodically. When it runs out of bricks, it will still be supplemented after informing someone. The robot itself cannot identify how many robots are onsite given the behavior of workers. In other words, the sub-collaboration system with only one robot and one BS worker in complex scenarios does not have essential differences comparing to the simple scenario, which can be simulated by the high-fidelity model. The only three minor differences are: 1) The robot will experience a larger check interval since the BS worker needs to deal with multiple robots simultaneously. In other words, the interval of robots gets checked may be longer than the check interval of the BS worker regulated by the supplement strategy; 2) When the BS worker receives the message from the EMR worker, the BS worker is possibly supplementing bricks for another robot, thus the BS worker may not immediately respond to the request; 3) As a result, the robot may not restart operating until the next check of the BS worker. To describe the differences, the modified flowchart of the BS worker is given in Figure 14. A new path (the red dotted rectangle) is used to describe the situation that the BS worker cannot respond immediately to the robot. We introduce three extra parameters to the high-fidelity simulation model. The interval of robots gets checked is called Get Checked Interval (GCI). GCI will completely replace CI in the high-fidelity model. Besides, Occupied Rate (OR) refers to the probability that the BS worker cannot respond immediately. If that happens, the robot needs to wait for Waiting Time (WT) before it is supplemented. These three parameters capture the inner relationship between complex scenarios and simple scenarios (summarized in Table 9). After the parameters are extracted, the values are assigned to the corresponding parameters in the modified simulation model. Therefore, we can simulate the sub-collaboration system by the high-fidelity model. Because all the robots are the same, the sub-collaboration system is essentially the average behavior of all the robots onsite. To this end, the whole complex scenarios are simulated. Note that simulating the complex scenario by averagely modeling its subparts will induce some error. However, considering it brings so many benefits as mentioned before, it is reasonable to sacrifice accuracy to some degree in return for better convenience and flexibility in the modeling process.

It is clear that this approach is suitable for the Multiple Robots-Multiple BS Workers scenario. Besides, the simple scenario with one robot and one BS worker can also be modeled after assigning CI to GCI and assigning 0 to OP and WT.

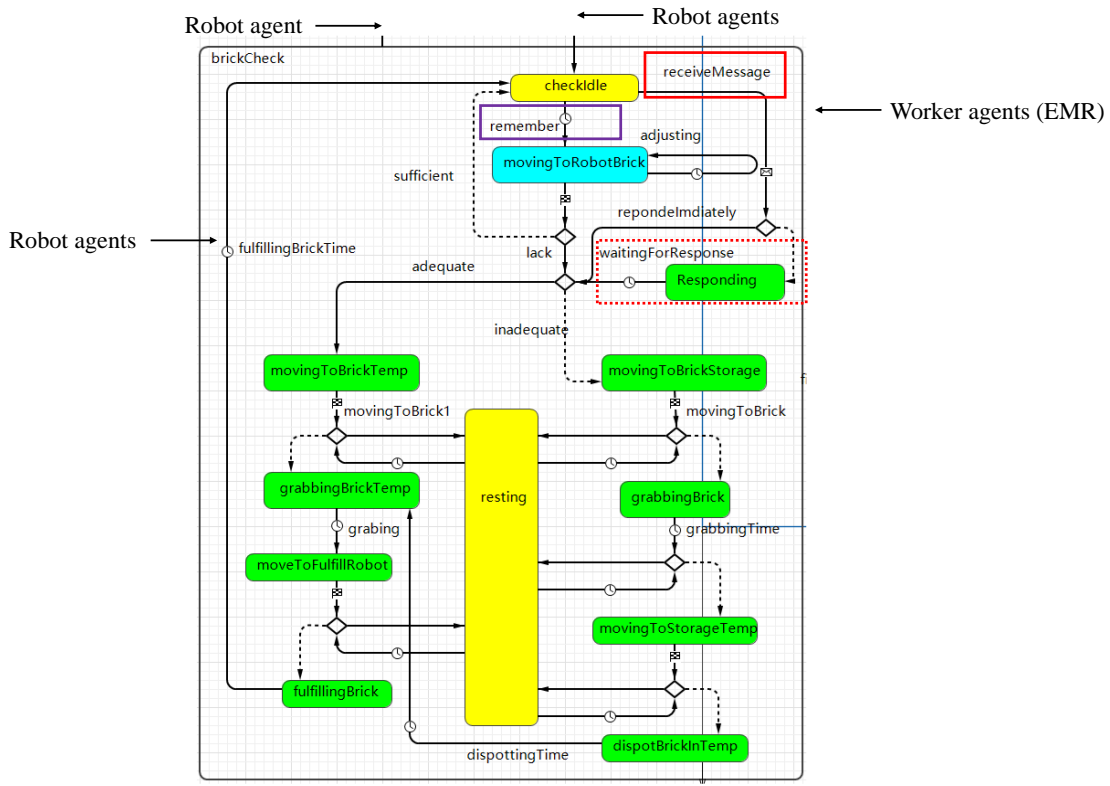

Figure 14. Modified flowchart of the BS worker for complex scenarios

Table 9. Key parameters capturing the relationship between complex and simple scenarios

| Parameter | Meaning |
| --- | --- |
| Get Checked Interval (GCI) | The check interval experienced by the robots |
| Occupied Rate (OR) | The probability that the robot cannot respond immediately |
| Waiting Time (WT) | The time the robot needs to wait before the BS worker give responses |

### 5.2.2 Extracting key parameters by the low-fidelity model

In order to model the complex scenarios with the simulation model, it is crucial to extract the value of GCI, OR, WT in these scenarios. A low-fidelity model is proposed to extract the parameters without the necessity to build a high-fidelity model for each complex scenario, as mentioned before. There are four main differences between the high-fidelity and low-fidelity model (Table 10).

- Platform: The high-fidelity model uses Anylogic© as the simulation platform, while the low-fidelity model is programmed using Object-Oriented-Programming in Python.

- Position and movement: The high-fidelity model considers the real position, physical movement and detailed behaviors of agents in the simulation environment. For example, when the BS worker starts to supplements the bricks for the robot, he needs to move to the storage place first, and then performs other actions such as grabbing and moving bricks. Considering the duration for each action is well defined and a task contains a series of actions, the duration of a task is the sum of durations for each action. For example, the time needed for a BS task contains the duration of moving, grabbing bricks, etc. Since the start position and end position are not fixed, the duration is a random variable. We collect the average duration of the whole checking process $T_c$, the supplementing process from Temp Storage Place $T_{st}$ and Long-term Storage Place $T_{sl}$ in the high-fidelity model. In the low-fidelity model, we ignore the position and movement of agents and directly use $T_c$, $T_{st}$ and $T_{sl}$ as the time needed for each corresponding task.

- Ergonomic behaviors: In the high-fidelity model, we consider the forgetting behavior and the fatigue of human workers. Besides, the triangular distributions are applied to capture workers' random performance. In the low-fidelity model, all these ergonomic behaviors are ignored to simplify the model.

- Flexibility: Due to the aforementioned three points, the low-fidelity model becomes much easier to build and modify than the high-fidelity model.

Table 10. Comparison between the high-fidelity and low-fidelity model

| The high-fidelity model | The low-fidelity model |
| --- | --- |
| Use Anylogic© as the simulation platform | Use Object-Oriented-Programming (OOP) in Python |
| Consider the position and movement of agents | Ignore the position and movement |
| Consider the ergonomic behaviors of workers | Ignore the ergonomic behaviors |
| Hard to build and modify | Easy to build and modify |

In the low-fidelity model, if the robot runs out of bricks and informs the BS workers to start the supplement process but all the BS workers are busy, it will be recorded as a failed request. Otherwise, if there is at least one available worker responds, it will be recorded as a successful request. For the scenario with gravity sensors, the alert that at least one BS worker can respond to is considered a successful request. Otherwise, the alert belongs to a failed request. Occupied Rate (OR) is calculated according to equation (4). $num_f$ and $num_s$ refer to the number of failed requests and successful requests respectively. When the robot is supplemented after a failed request, the model will calculate and record the interval between the request and the current time. Waiting Time (WT) is the average interval.

$$OR = \frac{num_f}{num_s + num_f} \qquad (4)$$

According to Figure 11, the definition of Check Interval (CI) is the interval between the time when the worker returns to "CheckIdle" state and the time the worker starts to check. The extraction of Get Checked Interval (GCI) in the robot's perspective is given in Figure 15 following the similar definition. For the scenario with two robot agents and one worker agent, GCI is longer than CI. The low-fidelity model will keep updating the latest end time of checking or supplementing. When the worker starts to check a robot, the model will calculate the interval between the latest end time and the current time. The interval is recorded as GCI for the corresponding robot. The average of all GCI for all robots will be calculated and used as the final value.

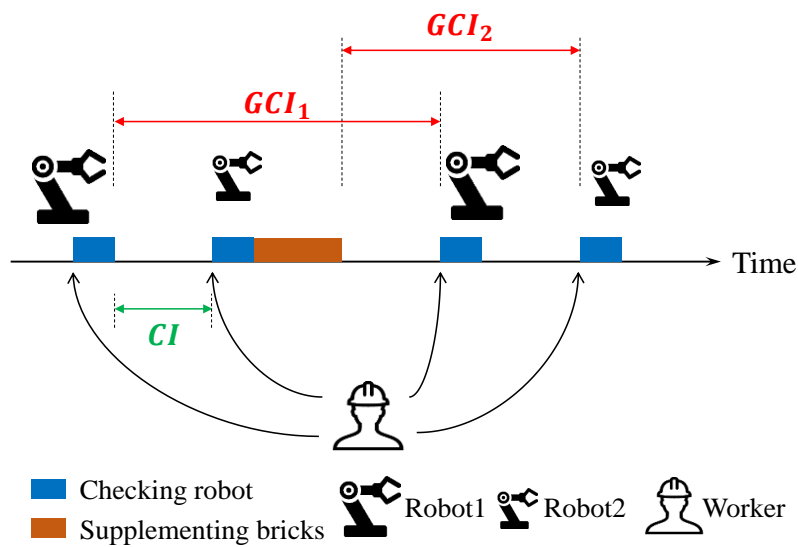

Figure 15. Check Interval (CI) in the robot's perspective

Then, the extracted OR, WT and GCI will be used to value the corresponding parameter in the high-fidelity model to simulate the sub-collaboration system in the complicated scenarios. The overall procedures are summarized in Figure 16.

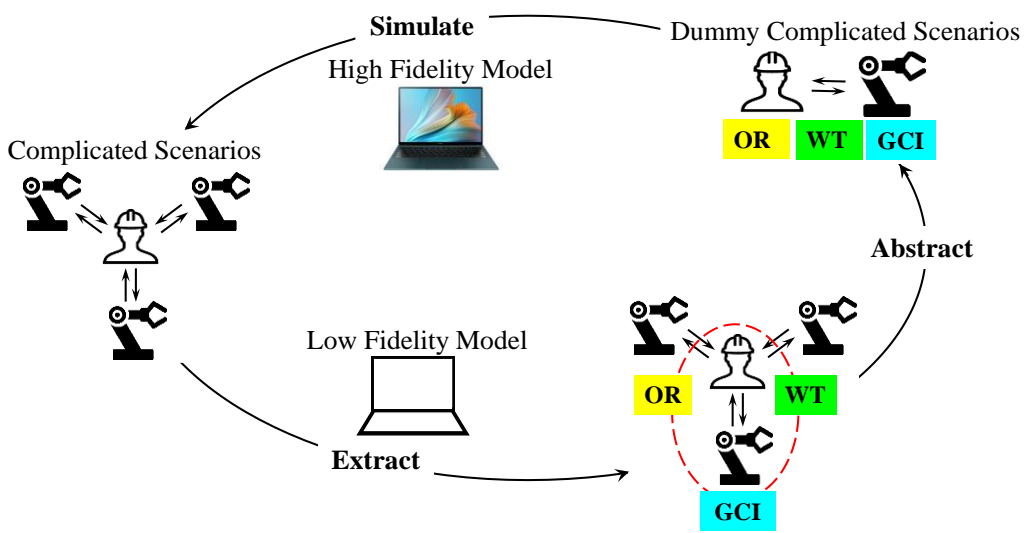

Figure 16. Overall procedures of scenario generalization

# 6 Simulation experiments

This section formulates simulation experiments for in-depth investigations into the influence of HRC on construction productivity. The influence is twofold. On the human side, workers can develop strategies to better collaborate with robots and improve construction productivity. Specifically, in the BS task, a better supplement strategy can be developed to reduce the downtime of the robots due to the lack of bricks. On the robot side, the robot can be designed to proactively interact with human workers. An example of the proactive design is to install gravity sensors. This section examines the twofold influence separately. Before that, the parameters in the model are calibrated, and the validation of the model is introduced.

## 6.1 Data collecting

Values of parameters involved in the model are given in Table 11 based on theoretical evidence, empirical evidence and necessary assumptions[35, 47]. More specifically, the third and the fifth parameters are directly observed from the video records[48]. Parameters related to the walking speed[49] and the forgetting behaviors[19, 44, 45] are derived from related researches. Note that the walking speed in [49] is around 1.3m/s. However, we consider that masons are under high workload and they usually need to walk in disordered and dangerous construction sites, so the value for Walking Velocity is smaller. The workers on site are usually experienced, therefore A and B equals 1 and 0.01[19]. The other parameters are based on assumptions. However, note that these parameters depend on the specific characteristics of robots and workers. The goal of this research is to propose a multi-fidelity modeling approach to simulate and analyze the HRC process without dealing with specific robots and workers. Therefore, making some assumptions is reasonable here. Also note that the value for these parameters can be easily adjusted when used in a specific circumstance. The worker agent also has two human-determined parameters, SL and CI. These parameters are determined by the supplement strategy, whose influence on productivity will be analyzed in this section.

As mentioned in Section 4.4, triangular distributions with a 20% variance are applied to the duration of preparation, installation, mortar removing, grabbing, dropping, and supplementing bricks. During each simulation experiment, AnyLogic© will determine the value of parameters according to the predefined constants or distributions in Table 11 to ensure that the parameters will not change in the collaboration process. The values of SL and CI are changed manually in different simulation scenarios.

Table 11. Value of parameters

| Agent | Index | Attribute | Value | Data source |
|---|---|---|---|---|
| Robot | 1 | Moving Velocity in Working | 0.2m/s | Assumptions (depends on the robot's type) |
| | 2 | Robot Storage Capacity | 300 bricks | |
| | 3 | Brick Laying Duration | 10s | Video records[48] |
| Worker | 4 | Walking Velocity | 0.75m/s | Previous research [49] |
| | 5 | Mortar Removing Duration | 5s | Video records[48] |
| | 6 | Supplement Duration | 10s | Assumptions (depends on the robot's type and the worker's characteristics) |
| | 7 | Preparation Duration | 20min | |
| | 8 | Installation Duration | 20min | |
| | 9 | Grabbing Duration | 30s | |
| | 10 | Dropping Duration | 30s | |
| | 11 | Moving Robot Velocity | 0.33m/s | |
| | 12 | Carrying Velocity | 0.67m/s | |
| | 13 | A | 1 | Previous research[19, 44, 45] |
| | 14 | B | 0.01 | |
| | 15 | SL and CI | Varying | Decided by the supplement strategy |

## 6.2 The randomness of experiment results

This section discusses the randomness in the experiments. We take three experiments as examples for better illustration. The values of SL and CI for these experiments are given in Table 12. Note that other parameters will have similar results. Each experiment is run five times. The construction time of each experiment is shown in Figure 17. For the first experiment, the range of construction time is from 392.53 hours to 392.58 hours. There is no big variation in the results. On contrary, for the other two experiments, we can see remarkable fluctuation in the construction time. The difference can reach to near 20 hours. The randomness is due to the triangular distributions with a 20% variance in the parameters mentioned in Section 4.4, which induce uncertainty to the system. Sometimes the uncertainty will not have much influence on the construction time (like the first experiment). However, note that the working hour agent restricts that the workers and robots can only work up to eight hours. If the work for the last day still needs nearly eight hours, another day may be needed to finish all the work due to a slight delay, inducing a remarkable change to the construction time.

Table 12. Experiment design to illustrate randomness

| Index | Parameters |
|---|---|
| 1 | SL = 300, CI=100, without sensors |
| 2 | SL = 300, CI=1800, without sensors |
| 3 | SL = 300, CI=3800, without sensors |

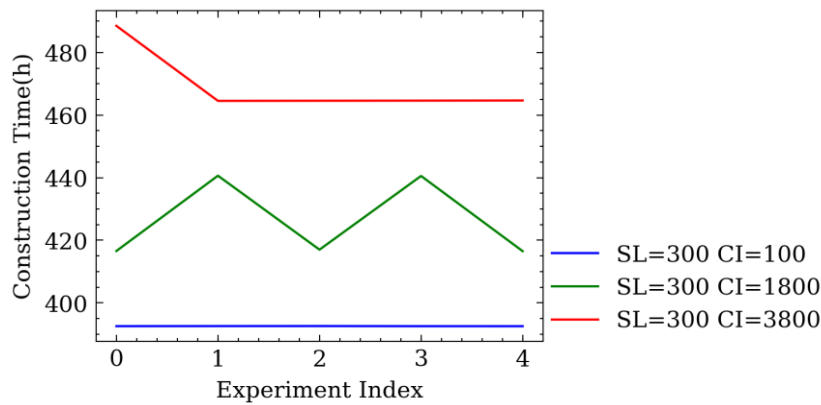

Figure 17. Experiment results to illustrate the randomness

Since the experiment has randomness, it is important to run each simulation experiment in the following sections five times to eliminate the effect of randomness. Then, the average construction time is selected as the measurement of construction productivity.

### 6.3 Model validation

The first experiment is the simple scenario with one robot and one BS worker. This setting is consistent with the real working scenario of SAM100 from video records[48]. Because it is the simple scenario, OR and WT equal 0. From the video record, it can be seen that the BS worker always stays with the robot, which reflects checking and supplementing of high frequency. Therefore, in this experiment, the value of SL and CI are 300 and 100, which represents frequent supplementing and checking respectively.

When SL equals 300 and CI equals 100, the total construction time is 392.55 hours. The construction task contains 38920 bricks in all, so averagely the robot agent can lay 2380 bricks per day. The comparison between the model results and the performance of real SAM100 reported in [10] is listed in Table 13. Results show that the model can accurately predict the performance of bricklaying robots, which shows superior applicability of the model. The remaining difference between the model result and the actual value is probably because the CI in the real-world application is smaller than 100. The inevitable simplification in the simulation, the error in the value selection of parameters, and the difference between the layout in the simulation environment and real-world application may be other reasons.

Table 13. Model validation

| Data source | Model result | SAM100[10] |
| --- | --- | --- |
| Bricks laid per day | 2380 | 2000 - 3000 |

## 6.4 Influence of supplement strategy

Supplement strategy is a key interaction between workers and robots, which determines the checking and supplementing frequency of BS workers. With a better strategy, BS workers can supplement bricks in time to avoid the breakdown of the robot. Therefore, it is important to study how supplement strategy influences construction productivity. The supplement strategy contains two parts. The first part is how a single BS worker collaborates with robots. Parameters related to this part are CI and SL. On the other hand, in the global view, the supplement strategy also determines the ratio between robots and BS workers. Since more BS workers will naturally make the checking more frequent. Note that gravity sensors are removed to emphasize the impact of the supplement strategy.

The scenario design is provided in Table 14. The first two scenarios focus on the supplement strategy of a single BS worker. For the Single robot –Single BS worker scenario, OR and WT both equal 0, and GCI equals CI. For the other two scenarios, we consider two robots and two BS workers at most in the construction site. Key parameters, OR, WT and GCI are extracted from the low-fidelity model. For each scenario, 60 groups of simulation experiments are conducted with different CI and SL according to Table 15. Since the maximum number of bricks in the robot is 300, the maximum value of SL is 300. Results from pretests show that 3800 seconds is enough to show the trend clearly, therefore the maximum value of CI is chosen as 3800. We consider 50 and 100 are small enough comparing to the maximum value, 300 and 3800. Thus, 50 and 100 are chosen to be the minimum values of SL and CI, respectively.

Table 14. Scenario design regarding the supplement strategy

| Index | Strategy | Work scenario |
|---|---|---|
| 1 | Supplement strategy of a single BS worker | Single robot –Single BS worker |
| 2 | | Multiple robots –Single BS worker |
| 3 | Human-robot ratio | Multiple robots –Multiple BS workers |

Table 15. Values of parameters related to supplement strategy

| Parameter | Range |
|---|---|
| CI (seconds) | 100, 200, 300, 800, 1300, 1800, 2300, 2800, 3300, 3800 |
| SL | 50, 100, 150, 200, 250, 300 |

### 6.4.1 The Single Robot-Single BS worker scenario

For the first scenario with one robot and one BS worker, construction time of different values of SL and CI are provided in Table 18 in Appendix A.

The overall effect of CI and SL on construction time is provided in Figure 19(a) using a contour map. Note that in this scenario, increasing SL and decreasing CI can reduce the construction time. This conclusion is straightforward. The BS worker only needs to focus on one robot, so larger SL and smaller CI mean more frequent checking and supplementing. However, this will also lead to the higher workload of the BS worker. Managers can use the proposed results to better balance the workload and construction productivity. Generally, the system can achieve good performance when SL is between 200 and 300 while CI is lower than 1500, which refers to the dark blue rectangle area in the figure.

As for the variation trend, the relationship between construction time and CI is given in Figure 20(a) and (b). All curves follow a rising trend. We track the GCI in the low-fidelity model of this scenario. When CI reaches 3800, for all SL from 50 to 300, the GCI is "infinite". To explain this phenomenon, remember that in our model we consider two ways to supplement bricks for robots. First, the BS worker will periodically check the bricks in the robot, and start the supplement process when bricks are less than SL. Second, if the robot has used up all the bricks and stopped operating, the EMR worker beside the robot will notice and inform the BS worker. Note that the robot will stop working for a while in the second way, which is harmful to the construction productivity. In this situation when CI reaches 3800, the interval is so long that the BS worker is not possible to check the robot in time. As a result, all the supplement process starts with the information from the EMR worker, which leads to the low productivity. This is the reason why 3800 is chosen as the upper bound of CI in the experiment design since larger CI will generate similar results. Besides, note that for smaller SL, the productivity curve will rise earlier. For example, when SL equals 50, the construction time will go beyond 460 hours when CI is just 800. The low-fidelity model indicates that this early rise is due to the impropriate ratio of CI and SL. For example, when SL equals 50 and CI equals 800, the CI is relatively much bigger than SL. As a result, although sometimes the bricks are more than SL, these bricks cannot sustain for the next interval. Similarly, when SL equals 100 and CI equals 1800, no supplement process follows the check process of the BS worker, so that the robot cannot avoid running out of bricks. The analysis indicates another important function of the low-fidelity model. It can help one to understand the simulation results of the high-fidelity model.

### 6.4.2 The Multiple Robots-Single BS Worker scenario

For the second scenario with two robots and one BS worker scenario, note that the variation trend is different. Despite the similar dark blue area for small CI and large SL, there is another dark read area corresponding to small CI and small CI in Figure 19(b). When SL is less than 100, CI is less than 500, the construction time is extremely high. It can be identified clearly in Figure 20 (c) and (d). The effect of CI on productivity reflects two types of variation trends. For larger SL (200, 250,300), the variation trend is similar to the first scenario. However, for smaller SL (50, 100, 150), the construction time first decreases and then goes up with growing CI. This means that in some circumstances, diligence (frequent checking) may have a counter effect on productivity.

To explain this counterintuitive trend, note that for scenarios with smaller SL, the productivity is more sensitive to the effectiveness of each check. Because when the SL is small, the BS worker does not start the supplement process until the bricks in the robot are relatively less. If one check is delayed and the robot only has a few bricks, these bricks may not sustain the next cycle. Also note that two robots are competing for the limited time of one BS worker in this scenario. Therefore, the supplement process for one robot may lead to the check for the other robot being delayed, which is likely to cause the other robots to run out of bricks. Besides, since all the supplement tasks are compacted to a small period, the Occupied Rate will be very high in this period, which makes the worker cannot respond immediately to the stop of robots. In contrast, when the SL is larger, the BS worker starts the supplement process early, thus the BS worker still has plenty of chances to check and supplement the robot in time before the bricks are used up.

The construction time data is provided in Table 19 in Appendix A.

### 6.4.3 The Multiple Robots-Multiple BS Workers scenario

For the third experiment with two robots and two BS workers, the general distribution of construction time is similar to the single robot-single BS worker scenario, as shown in Figure 19. The intense competition for the time of BS workers is relieved by introducing extra labor forces. As a result, the dark red area in Figure 19(b) dissipates. Note that the dark blue area is similar to the Single Robot-Single BS Worker scenario, but some green area is replaced by the blue area in Figure 19(c). This indicates that productivity increases although the ratio between robots and workers remains the same. This implies that there exists a scale effect between the collaboration of robots and BS workers, especially for scenarios with larger CI. The reason behind this is that the large portion of overlap between check intervals makes that the GCI smaller than the CI determined by the supplement strategy (Figure 18).

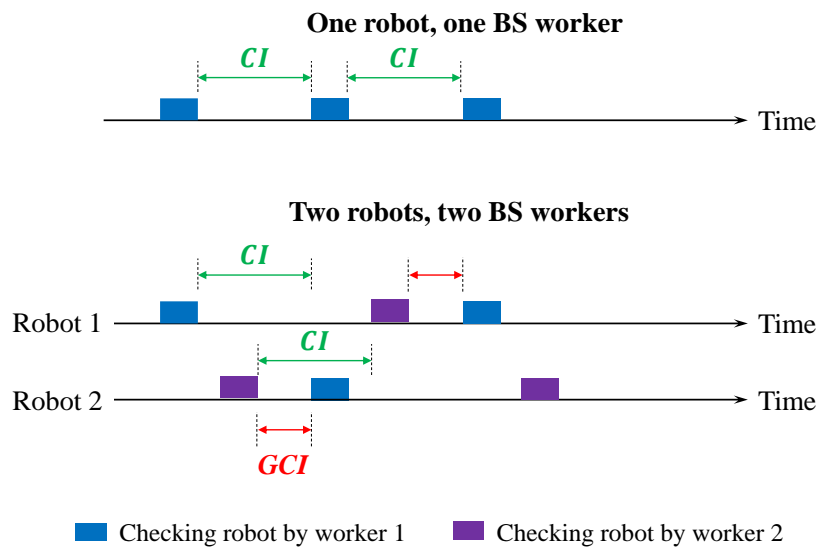

Figure 18. Explanation of the scale effect

The construction time data is provided in Table 20 in Appendix A.

### 6.5 Influence of proactive interaction

This section investigates how proactive interaction designs, namely the installation of gravity sensors, affect construction productivity. Gravity sensors are used in the simulation. Note that the values of CI and SL still vary according to Table 15. The only difference is that 300 is removed from the possible value of SL. Otherwise, the sensor will send alerts just after the bricks are supplemented, which is not realistic.

#### 6.5.1 The Single Robot-Single BS Worker scenario

The construction time distribution with SL and CI is provided in Table 21 and Figure 19(d). Note that CI, which has a great influence on the productivity in previous sections, does not have much impact. This is because the BS workers do not need to rely on checking to obtain the information thanks to the gravity sensors. Therefore, with the sensors installed, workers do not need to check frequently. The workers' burden is greatly relieved. Furthermore, the decrease of construction time, namely the improvement of construction productivity, is provided in Figure 21(a). Results show that the sensor can always improve productivity regardless of the value of CI and SL. Averagely, productivity can improve by 4.94%. Specifically, the improvement is remarkable when SL and CI are larger. According to our experiments, when SL and CI are greater than 200 and 2800 respectively, the productivity can generally improve more than 10%. Then improvement is up to 18.04% when SL and CI equal 250 and 3800 respectively. As discussed before, large SL refers to potentially frequent supplementing. But the supplement strategy with large CI does not fully exploit the potential when there are no gravity sensors. The installation of gravity sensors eliminates the impact of CI and helps these scenarios to achieve their full potential. On contrary, when CI is less or equal to 300, the improvement is neglectable. This phenomenon is due to the overlapping effect of gravity sensors and periodical checks. Note that if CI is relatively small, the BS workers are able to identify the time when the bricks are less than SL and start the supplement process. In this circumstance, the existence of gravity sensors can only move the check a little ahead of schedule, which does not have much influence on the overall performance. From this, we conclude that directly introducing gravity sensors may not be a good idea since sensors are expensive. The adoption of gravity sensors needs to coordinate with the development of supplement strategy to fully exploit the benefits of sensors. To this end, the twofold influence, namely the supplement strategy and the design for proactive interaction, are connected.

### 6.5.2  The Multiple Robots-Single BS Worker scenario

For the Multiple Robots-Single BS worker scenario, some interesting observations are listed below.

1. According to Figure 19(e), CI does not have much influence on construction productivity. Similar to the previous scenario, the existence of gravity sensors eliminates the effect of CI.

2. The productivity improvement is provided in Figure 21(b). It can be seen that the gravity sensors can always improve construction productivity as well. Averagely, the productivity can improve by 11.04%, which is more effective than the Single Robot-Single BS Worker scenario. This indicates that the human-robot ratio influences the improvement of productivity. When the human-robot ratio decreases, the average workload increases, which further needs the sensors to help coordinate. Therefore, the effect of gravity sensors is more remarkable when the human-robot ratio is smaller.

3. Specifically, different from the previous scenario, when CI and SL are both small, the construction productivity

has significant improvements. When SL and CI are less or equal to 100 and 200 respectively, the productivity can increase 17% to 22%. As discussed before, in this area, the competition between two robots will increase the Occupied Rate of the BS worker. Thanks to the extra information provided by the gravity sensors, the checking and supplementing process can be well organized which can relieve the intense work.

4. From the blue area in Figure 21(b), it can be seen that the supplement strategy can greatly affect the effect of introducing gravity sensors.

The construction time data is provided in Table 22 in Appendix A.

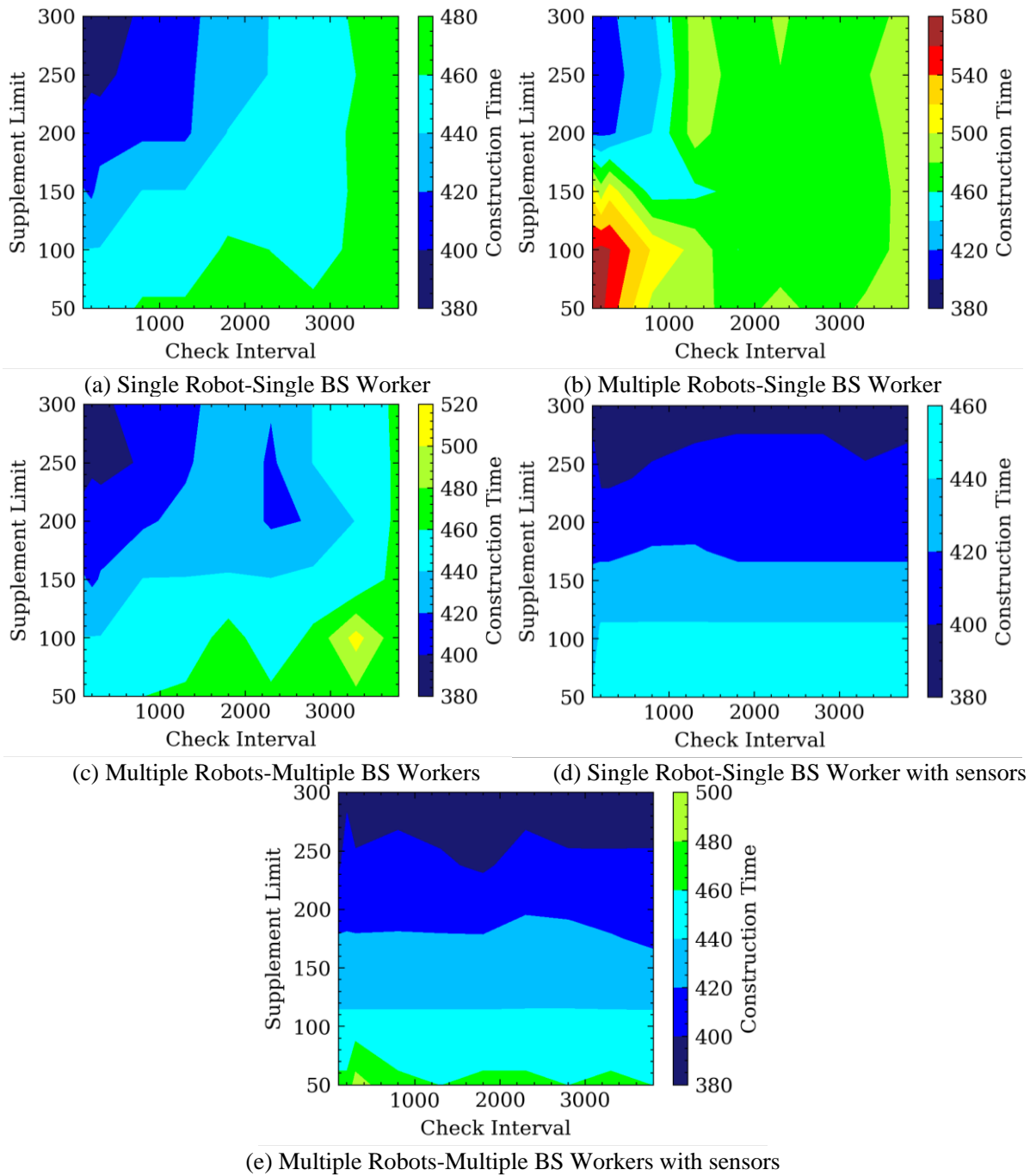

(a) Single Robot-Single BS Worker
(b) Multiple Robots-Single BS Worker
(c) Multiple Robots-Multiple BS Workers
(d) Single Robot-Single BS Worker with sensors
(e) Multiple Robots-Multiple BS Workers with sensors

Figure 19. Construction time for the different scenarios

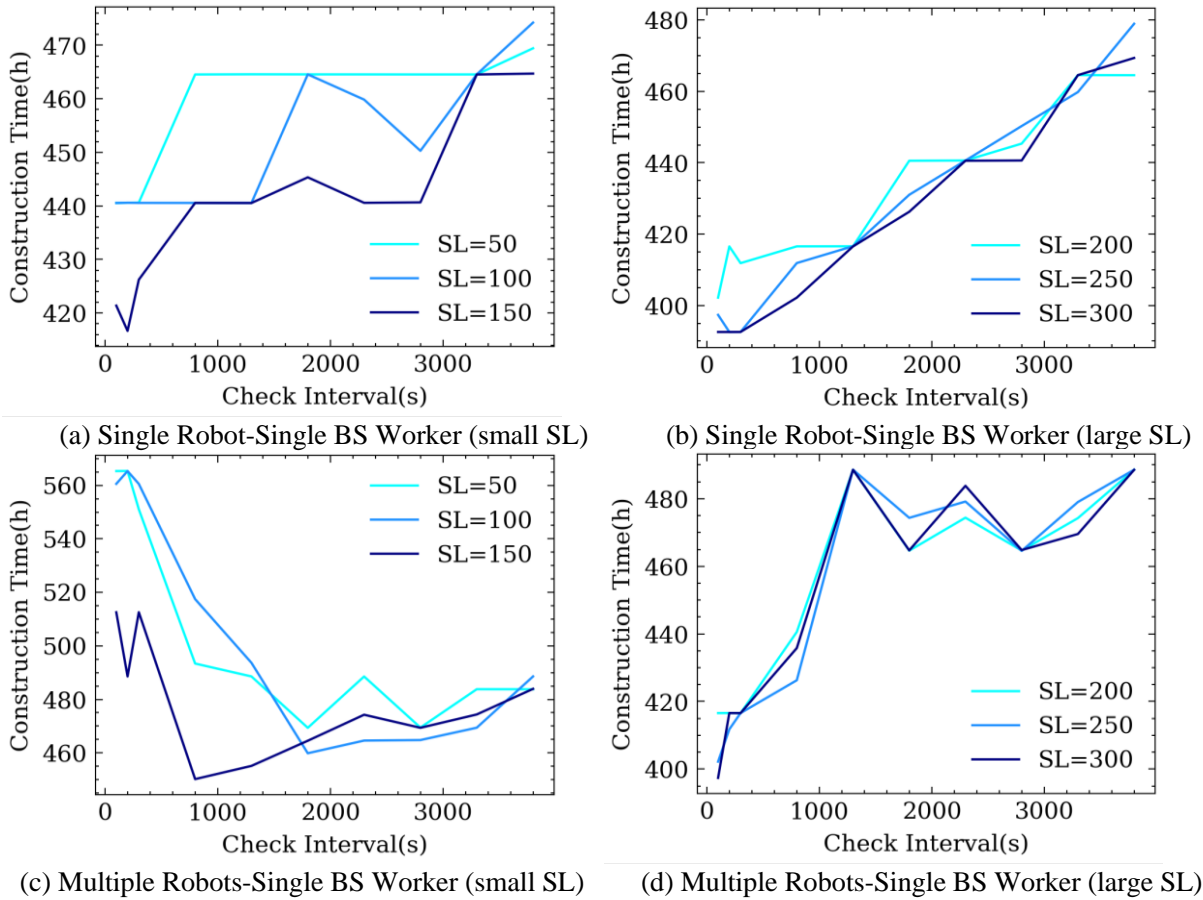

(a) Single Robot-Single BS Worker (small SL)  (b) Single Robot-Single BS Worker (large SL)

(c) Multiple Robots-Single BS Worker (small SL)  (d) Multiple Robots-Single BS Worker (large SL)

Figure 20. The effect of CI on productivity for multiple robots-single BS worker scenarios

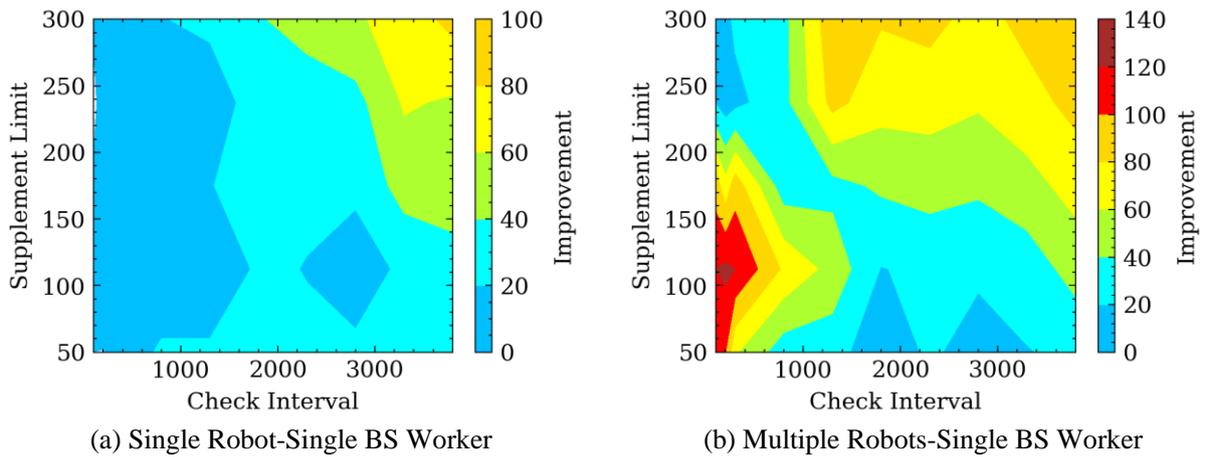

(a) Single Robot-Single BS Worker  (b) Multiple Robots-Single BS Worker

Figure 21. The improvement of proactive interaction

# 7  Discussion

In this section, we review the whole paper and discuss the key points in the development, utilization and potentials of the simulation model.

## 7.1 Human-robot collaboration characteristics

In Section 2.1, we discuss that the complexity and uniqueness in the HRC process in construction are because 1) it simultaneously involving multiple communication and interaction modes; 2) it stresses safety requirements; 3) humans and robots show different characteristics in collaboration. All three factors are well considered in the simulation model, which shows the great applicability of our model in the HRC process in construction.

- Table 16 and Table 17 show the communication modes and interaction modes related to the bricklaying process. It shows that HRC process simultaneously involves three communication modes and three interaction modes. These modes are successfully captured by the proposed model.

Table 16. Involved communication modes in the AB model

| Communication mode | Scenario |
| --- | --- |
| Direct physical interaction | The BS worker supplements bricks for the robot; Workers move the robot; |
| Remote contactless interaction | The Robot waits for EMR for the next layer; Sensors alert workers; |
| Message exchange | Workers install the robot; |

Table 17. Involved interaction modes in the AB model

| Interaction mode | Scenario |
| --- | --- |
| Coexistence | BS tasks and robot laying bricks happen at the same time, same place but different items and tasks |
| Cooperation | The EMR worker and the robot work at the same time, same place, same item (a wall), but different tasks |
| Collaboration | The BS task is performed by the BS worker and the robot |

- As mentioned before, the robot will not start laying the next layer until the EMR worker finishes the layer. Besides, the minimum distance between robots and workers is required. These two rules represent the safety requirements. Other requirements can be embedded in the model as well in the same way.

- The different characteristics of humans and robots can be easily modeled by determining different behavior rules for two agents thanks to the nature of AB simulation. In this research, forgetting, fatigue and behavior uncertainty are considered.

## 7.2 Inspiration from the simulation results

From the simulation results, the influential factors have a complicated, and profound influence on construction productivity. Generally, it is highly recommended to adopt a robot design that can proactively interact with workers (i.e., adopting gravity sensors) and hire more workers. These approaches are proved to be always beneficial to construction productivity according to the simulation. Furthermore, note that in scenarios with multiple robots, although the average performance regarding one robot is inferior to the performance of the scenario with a single robot (Figure 19), the existence of multiple robots can multiply productivity. Hence, introducing more robots to construction can improve productivity as well. Considering all three measurements (introducing gravity sensors, more workers and robots) will increase the cost, there is essentially a trade-off between productivity and cost. We want to stress that the proposed simulation model can accurately examine the benefits of each measurement, thus it is a powerful tool for managers in estimating the trade-off.

There are also ways to improve productivity without introducing many costs. We also prove that a better supplement strategy, namely adopting proper values for SL and CI, can greatly improve productivity. The manager can test and choose a good supplement strategy by adjusting the parameters in the proposed simulation model, or building their model following the key modeling points in Section 4 and 5. The existence of the low-fidelity model makes the process very easy, since the majority of work in revising and building models is on the low-fidelity model instead of having to build their own high-fidelity model.

## 7.3 Future model development and analysis

In this research, we only focus on the influence of the HRC process on productivity as one aspect of construction performance. Specifically, major attention is paid to the influence of the collaboration strategy of workers and the proactive design of robots. Admittedly, there are still some important issues we can add and further enrich the model. Some other influencing factors can also be analyzed based on the simulation model. However, since the space is limited, we cannot cover all the aspects. This section discusses the future directions of this research.

### 7.3.1 Influence of human factors

Three human factors, forgetting behaviors, muscle fatigue and behavior uncertainty, highlight the difference between humans and robots. It will be interesting to investigate the influence of the three parameters. We plan to design some experiments to analyze the variation trend while adjusting the related parameters. For example, we can magnify the values in Table 7 or A, B to represent workers that are more easily to feel tired or forget, respectively. The challenge is to pick the values of these parameters in a scientifically rigid way.

#### 7.3.2 Influence of safety issues

Besides, the safety issue is also very important for the HRC process. In this research, we mainly consider the two safety requirements. (1) Robots will not start working on the next layer until the EMR worker finishes the current layer. (2) Robots will stop working when there is a worker nearby. In the future, we plan to consider more safety requirements to make the simulation more realistic. When there are more safety rules, it is more convenient to formulate a new abstract safety agent that takes charge of all safety requirements. With this safety agent, it will be easy to add, delete or modify safety rules since all the rules are under centralized management. Furthermore, we can investigate the influence of the safety issue. For example, we can count how many times the robot stops due to a nearby worker, and then explore how to optimize the site layout to reduce the stop time.

#### 7.3.3 Influence of construction quality

The construction quality issue is also worthy of investigation. We plan to consider the random error of the position of newly generated brick agents in future work. The challenge of this approach is to correlate the construction quality with the HRC process. One approach is to consider that this position error may be accumulated due to the possible initial installation error. To avoid the accumulated error, it is interesting to design a new supervision task for worker agents. The new worker will stop and reinstall the robot if the accumulated error is noticeable.

## 8 Conclusion

Considering the complexity and uniqueness of HRC process in construction, it is still unclear how it will impact construction productivity. In this research, an agent-based multi-fidelity simulation approach is applied to address this question. The following conclusions are drawn from this study.

1. The proposed simulation method shows accuracy, applicability and flexibility. On one hand, the simulation model can accurately predict the performance of the bricklaying process and cover multiple collaboration modes and interaction modes, which shows superior applicability of the model. On the other hand, the utilization of the multi-fidelity simulation approach makes the model easy to be revised and applied to other scenarios. Besides, the simulation model can help examine the effects of possible optimization solutions virtually instead of having to test them on the spot, which can reduce the cost and risk.

2. The supplement strategy has a great influence on productivity. Generally, when there is only one robot, higher Supplement Limit (SL) and smaller Check Interval (CI) can improve construction productivity. However, when there are more robots onsite but the workers remain the same, the competition between robots for the limited time of workers will worsen the construction performance, especially when SL and CI are both small.

3. The research also reveals the scale effect in the collaboration system. If there are more robots and workers, even if the human-robot ratio remains the same, the productivity per robot can be improved.

4. The installation of gravity sensors can enable robots to proactively interact with workers. Introducing sensors can improve construction productivity significantly, up to 22%. The research also finds that the installation of sensors may not always have a remarkable effect. The effect is influenced by the supplement strategy and the human-robot ratio. The effect is generally more remarkable when SL and CI are large and when the human-robot ratio is small.

Overall, this research contributes an integrated approach to simulate and evaluate HRC's impacts on construction productivity as well as valuable insights on how to optimize HRC for better performance. The authors summarize key points in simulation and show a paradigm concerning analyzing the effect of HRC based on the model, which allows this method to be extended to other robot-aided construction domains besides bricklaying. The proposed approach is also useful for the development of new robots, optimization of HRC process to maximize construction performance and occupational health.

Further research can be conducted to better develop and utilize the proposed model. Many parameter values in this research are not based on actual robots due to the limited practical application. As construction robots become more prevalent, parameters can be adjusted to actual values. Besides, as discussed in Section 7, it is interesting to discuss the influence of human factors, safety issues and construction quality based on the simulation model. Finally, this research focuses on the construction domain. Using the proposed modeling approach to simulate and analyze the human-robot collaboration process in other industries will also be meaningful.

## Acknowledgment

The research is supported by the National Key R&D Program of China (No. 2018YFD1100900), the National Natural Science Foundation of China (No. 51908323), the Tsinghua University Initiative Scientific Research Program (No. 2019Z02UOT) and Tsinghua University Students Research Training Program.

## Declaration of competing interest

The authors declare that they have no known competing financial interests or personal relationships that could have appeared to influence the work reported in this paper.

# Appendix A. Simulation Results

Table 18. Construction time for the Single Robot-Single BS Worker Scenario

| SL/CI | 100 | 200 | 300 | 800 | 1300 | 1800 | 2300 | 2800 | 3300 | 3800 |
|---|---|---|---|---|---|---|---|---|---|---|
| 50 | 440.544 | 440.594 | 440.619 | 464.546 | 464.580 | 464.564 | 464.560 | 464.549 | 464.554 | 469.435 |
| 100 | 440.554 | 440.596 | 440.567 | 440.560 | 440.552 | 464.560 | 459.850 | 450.290 | 464.567 | 474.234 |
| 150 | 421.381 | 416.663 | 426.224 | 440.559 | 440.546 | 445.339 | 440.578 | 440.661 | 464.543 | 464.708 |
| 200 | 402.174 | 416.539 | 411.854 | 416.556 | 416.581 | 440.559 | 440.635 | 445.361 | 464.563 | 464.542 |
| 250 | 397.346 | 392.561 | 392.565 | 411.874 | 416.590 | 431.016 | 440.621 | 450.287 | 459.811 | 478.963 |
| 300 | 392.553 | 392.549 | 392.564 | 402.181 | 416.540 | 426.251 | 440.563 | 440.631 | 464.554 | 469.378 |

Table 19. Construction time for the Multiple Robot-Single BS Worker Scenario

| SL/CI | 100 | 200 | 300 | 800 | 1300 | 1800 | 2300 | 2800 | 3300 | 3800 |
|---|---|---|---|---|---|---|---|---|---|---|
| 50 | 565.33 | 565.38 | 551.05 | 493.39 | 488.56 | 469.36 | 488.54 | 469.46 | 483.79 | 483.77 |
| 100 | 560.57 | 565.34 | 560.53 | 517.47 | 493.65 | 459.85 | 464.60 | 464.79 | 469.40 | 488.57 |
| 150 | 512.55 | 488.57 | 512.58 | 450.19 | 455.13 | 464.53 | 474.25 | 469.36 | 474.35 | 483.93 |
| 200 | 416.56 | 416.58 | 416.55 | 440.56 | 488.58 | 464.63 | 474.39 | 464.66 | 474.26 | 488.58 |
| 250 | 402.19 | 411.76 | 416.57 | 426.28 | 488.60 | 474.34 | 479.15 | 464.56 | 479.02 | 488.58 |
| 300 | 397.37 | 416.57 | 416.56 | 435.79 | 488.57 | 464.71 | 483.81 | 464.81 | 469.55 | 488.54 |

Table 20. Construction time for the Multiple Robot- Multiple BS Worker Scenario

| SL/CI | 100 | 200 | 300 | 800 | 1300 | 1800 | 2300 | 2800 | 3300 | 3800 |
|---|---|---|---|---|---|---|---|---|---|---|
| 50 | 440.59 | 440.55 | 440.61 | 459.77 | 464.66 | 474.18 | 464.61 | 464.60 | 474.19 | 464.56 |
| 100 | 440.54 | 440.64 | 440.62 | 440.61 | 445.45 | 469.42 | 445.43 | 464.56 | 507.74 | 464.54 |
| 150 | 421.37 | 416.62 | 421.38 | 440.58 | 440.60 | 440.56 | 440.55 | 445.38 | 440.63 | 469.36 |
| 200 | 406.96 | 407.03 | 411.81 | 416.56 | 426.18 | 435.77 | 416.56 | 421.35 | 440.57 | 464.55 |
| 250 | 392.60 | 397.36 | 392.61 | 402.17 | 416.58 | 435.77 | 416.56 | 440.56 | 440.55 | 464.56 |
| 300 | 392.57 | 397.35 | 392.55 | 416.56 | 416.58 | 426.34 | 421.35 | 440.57 | 445.35 | 464.57 |

Table 21. Construction time for the simple scenario with gravity sensors

| SL/CI | 100 | 200 | 300 | 800 | 1300 | 1800 | 2300 | 2800 | 3300 | 3800 |
|---|---|---|---|---|---|---|---|---|---|---|
| 50 | 440.58 | 440.60 | 440.56 | 440.57 | 440.57 | 440.55 | 440.56 | 440.53 | 440.56 | 440.56 |
| 100 | 435.84 | 440.55 | 440.56 | 440.56 | 440.57 | 440.55 | 440.61 | 440.58 | 440.57 | 440.63 |
| 150 | 416.54 | 416.59 | 416.57 | 421.36 | 421.36 | 416.58 | 416.58 | 416.55 | 416.55 | 416.58 |
| 200 | 411.77 | 397.36 | 397.44 | 402.25 | 406.95 | 411.77 | 411.76 | 411.78 | 402.37 | 407.04 |
| 250 | 392.53 | 392.53 | 392.57 | 392.61 | 392.56 | 392.55 | 392.54 | 392.57 | 392.53 | 392.56 |

Table 22. Construction time for the complex scenario with gravity sensors

| SL/CI | 100 | 200 | 300 | 800 | 1300 | 1800 | 2300 | 2800 | 3300 | 3800 |
|---|---|---|---|---|---|---|---|---|---|---|
| 50 | 464.58 | 464.56 | 488.58 | 464.65 | 459.74 | 464.66 | 464.66 | 459.83 | 464.61 | 459.76 |
| 100 | 440.56 | 440.55 | 440.57 | 440.57 | 440.56 | 440.55 | 440.59 | 440.62 | 440.58 | 440.55 |
| 150 | 421.42 | 421.41 | 421.40 | 421.42 | 421.39 | 421.35 | 426.18 | 426.20 | 421.34 | 416.58 |
| 200 | 397.43 | 406.97 | 402.26 | 406.96 | 402.16 | 397.44 | 406.94 | 402.25 | 402.19 | 402.32 |
| 250 | 392.56 | 397.35 | 392.56 | 392.57 | 392.60 | 392.52 | 392.54 | 392.56 | 392.52 | 392.56 |